%% file: main.tex
\documentclass[lettersize,journal]{IEEEtran}

\newcommand{\R}{\mathbb{R}}

\usepackage{multirow, booktabs}
\usepackage[export]{adjustbox}
\usepackage[dvipsnames, table, xcdraw]{xcolor}
\newcommand{\revtextiot}[1]{{\color{black}#1}}
\definecolor{revcolor}{RGB}{0, 0, 0}
\newcommand{\revtext}[1]{{\color{black}#1}}
\newcommand{\revtextrewrite}[1]{{\color{black}#1}}

\usepackage{ifpdf}
\usepackage{cite}
\usepackage{amsmath}
\usepackage{amsfonts}
\usepackage{algorithmic}
\usepackage{array}
\usepackage{booktabs}       
\graphicspath{{figs/}}     
\usepackage[caption=false,font=footnotesize]{subfig}
\usepackage{url}
\usepackage{verbatim}
\usepackage{graphicx}
\usepackage{cite}
\usepackage{textcomp}
\usepackage{stfloats}

\begin{document}
\input{chaps/00-cov-abs/00-cover-ieee}
\maketitle

\input{chaps/00-cov-abs/00-abstract}
\input{chaps/01-intro/01-introduction}
\input{chaps/02-rw/02-related_work}
\input{chaps/03-fw/03-framework}

\input{chaps/04-doe/04-1-casestudies}

\input{chaps/04-doe/04-experiment}
\input{chaps/05-results/05-results}
\input{chaps/06-conclusion}

\section{Data and Code Availability}
The script we used to generate this synthetic dataset is available in the associated code repository to ensure reproducibility and facilitate further studies. Our code and data will be made available after acceptance of the
manuscript under \url{https://github.com/MengjieZhao/dyedgegat}.

\section*{Acknowledgments}
This work was supported by the Swiss National Science Foundation under Grant 200021\_200461.
ChatGPT has been used to correct the grammar of the text and for proofreading.

\bibliographystyle{IEEEtran}
\bibliography{references}

\input{chaps/08-biography}
\end{document}

%% file: chaps/00-cov-abs/00-cover-ieee.tex
%
\title{\textcolor{black}{DyEdgeGAT: Dynamic Edge via Graph Attention for Early Fault Detection in IIoT Systems}}   
%
%
%

\author{Mengjie~Zhao and
        Olga~Fink~\IEEEmembership{Memeber,~IEEE}
\thanks{The authors  with the Laboratory of Intelligent Maintenance and Operation
Systems, EPFL, 1015 Lausanne, Switzerland (e-mail: mengjie.zhao@epfl.ch, olga.fink@epfl.ch).}
}

%
%

\markboth{Dynamic Edge via Graph Attention for Early Fault Detection in IIoT Systems}%
{Shell \MakeLowercase{\textit{et al.}}: Bare Demo of IEEEtran.cls for IEEE Journals}
%

%% file: chaps/00-cov-abs/00-abstract.tex
\begin{abstract}
In the Industrial Internet of Things (IIoT), condition monitoring sensor signals from complex systems often exhibit nonlinear and stochastic spatial-temporal dynamics under varying conditions. These complex dynamics make fault detection particularly challenging.
While previous methods effectively model these dynamics, they often neglect the evolution of relationships between sensor signals. Undetected shifts in these relationships can lead to significant system failures. \revtextiot{Furthermore, these methods frequently misidentify novel operating conditions as faults.
Addressing these limitations, we propose DyEdgeGAT~\textit{(Dynamic Edge via Graph Attention}), a novel approach for early-stage fault detection in IIoT systems. 
DyEdgeGAT's primary innovation lies in a novel graph inference scheme for multivariate time series that tracks the evolution of relationships between time series, enabled by dynamic edge construction. 
Another key innovation of DyEdgeGAT is its ability to incorporate operating condition contexts into node dynamics modeling, enhancing its accuracy and robustness.
We rigorously evaluated DyEdgeGAT using both a synthetic dataset, simulating varying levels of fault severity, and a real-world industrial-scale multiphase flow facility benchmark with diverse fault types under varying operating conditions and detection complexities. 
The results show that DyEdgeGAT significantly outperforms other baseline methods in fault detection, particularly in the early stages with low severity, and exhibits robust performance under novel operating conditions.
}

\end{abstract}


\begin{IEEEkeywords}
Graph Neural Networks, Graph Learning, Multivariate Time Series, Unsupervised Fault Detection
\end{IEEEkeywords}

%% file: chaps/01-intro/01-introduction.tex
\section{Introduction}


\IEEEPARstart{T}{he} increasing deployment of sensors in industrial systems has enabled the collection of extensive Multivariate Time Series (MTS) data, facilitating the condition monitoring of complex systems to detect the onset of critical faults as early as possible~\cite{mohammadi2018deep}.
The complex dynamics of monitored systems, characterized by interconnected subsystems and components, often result in strong spatial-temporal dynamics of the heterogeneous MTS data \cite{younan2020challenges}~\cite {dong2023iotreview}. Due to the high interdependencies in the data, it becomes more challenging to detect incipient faults in such systems.

\revtext{
Effective fault detection is crucial for preventing severe system failures and improving system reliability. 
Particularly, detecting faults at their early stage can also contribute to extending the useful lifetime of components by preventing too early preventive replacements. Taking preemptive actions based on early fault detection can be instrumental in maintaining optimal system performance and avoiding costly disruptions.
While fault detection is relatively straightforward at higher severity levels when faults have fully manifested in the sensor measurements, identifying incipient faults, where they have not yet caused noticeable impacts on the system's performance, poses a greater challenge~\cite{wei2020robust}. 
The primary difficulty lies in detecting subtle changes, while an additional challenge arises from the risk of overly sensitive algorithms, potentially leading to an excessive number of false alarms~\cite{chao2019hybrid}.

Different types of faults, each with different difficulties to detect, can affect the system. These range from sensor faults that only impact one signal and are relatively straightforward to detect, to complex faults that affect multiple components. The latter often leads to secondary fault impacts and are considerably harder to detect~\cite{lughofer2019predictive}. Since faults are rare, recently unsupervised fault detection methods have been increasingly applied, influenced by advancements in anomaly detection research. One type of faults that current state-of-the-art methods have insufficiently addressed is the detection of changes in relationship between signals and components. Such faults can remain undetected for extended periods because the individual observations may appear to exhibit healthy conditions in terms of their functional behavior~\cite{boyes2018industrial}.}

\input{tabs/overview}

\revtext{
To effectively detect a change in these functional relationships, it is essential to accurately model these relationships.
Traditionally, modeling relationships between MTS in the IIoT context has mainly focused on two aspects: explicit functional relationship modeling and dynamics modeling.
Feedford Neural Networks (FNN)~\cite{ahmed2014automotive} and Autoencoders (AE)~\cite{guo2020unsupervised} have been successful in explicit functional relationship modeling. Recurrent Neural Networks (RNN)~\cite{ergen2019unsupervised} and Convolutional Neural Networks (CNN)~\cite{garcia2022temporal} have been successful at capturing system dynamics. However, these methods often fall short of addressing spatial-temporal dynamics within complex systems.
In response to these limitations, Graph Neural Networks (GNNs) have emerged as a promising alternative~\cite{wu2020comprehensive}\cite{jin2023survey}.
By establishing a graph structure from the MTS data, with each time series represented as a node and edges indicating interactions between different time series, GNNs can learn the spatial-temporal relationships between the sensors~\cite{zawislak2017graph}.
\revtextiot{Table~\ref{tab:intro_overview} provides a comparative analysis of different approaches in modeling these relationships.}
Several Spatial-Temporal GNN (STGNN)-based methods have been developed specifically for modeling spatial-temporal dynamics within IIoT systems.
For instance, Multivariate Time-series Anomaly Detection via Graph Attention Networks (MTADGAT)~\cite{zhao2020mtadgat} utilizes graph attention networks (GATs) to simultaneously construct a feature-oriented graph and a time-oriented graph, capturing both spatial and temporal dynamics. Additionally, Graph Deviation Network (GDN)~\cite{deng2021gdn} employs node embeddings to capture the unique characteristics of each sensor and utilizes an attention mechanism incorporating these sensor embeddings to better predict the future behavior of sensors, addressing the heterogeneity of sensor data in IIoT.

While these methods effectively model spatial-temporal dynamics within MTS, they generally overlook the evolution of functional relationships between system variables. This aspect is crucial for accurately detecting changes in relationships, which are often indicative of system faults. Current spatial-temporal GNNs are unable to capture these dynamic changes.
Tracking relationship shifts within IIoT systems to enable early fault detection poses two main challenges:
\begin{enumerate}
    \item \textbf{Static graph relation}. Existing STGNNs typically assume the relationship in the MTS data within a defined observation time period does not change. While these approaches can partially account for the dynamics within the data by constructing a new graph for a new observation window, they cannot capture the temporal evolution of relationships (edge weights) within the data.
    \item \textbf{Distinguishing faults from novel operating conditions}. Training data collected under healthy conditions may not contain all operational scenarios. Identifying whether relationship shifts in test time arise from faults or novel operating conditions remains challenging.
\end{enumerate}
}

\revtext{
To address the limitations outlined above, we introduce \textbf{DyEdgeGAT} (\textit{Dynamic Edge via Graph Attention}), a novel framework designed to capture relationship shifts in MTS data for effective early fault detection.
DyEdgeGAT addresses the challenge of static graph inference by dynamically inferring edges between time series, and constructing an aggregated temporal graph for MTS data. This enables us to capture not only \textbf{node dynamics} (i.e., \textit{the intrinsic dynamics of a sensor}) but also \textbf{edge dynamics} (i.e.,\textit{ the evolving relationships between sensors}). The term ``dynamic'' in DyEdgeGAT specifically refers to the evolving sequence of edge weights over time, reflecting the evolution of pairwise relationships between nodes in the temporal graph. This is contrary to the definition used in previous methods that define it as constructing a new static graph per observation window.  
\revtextiot{To address the challenge of distinguishing fault-induced relationship changes from novel operating conditions}, DyEdgeGAT innovatively differentiates between \textbf{system-dependent} variables (e.g., \textit{control variables} that are explicitly set and \textit{external factors} outside of the system) and \textbf{system-independent} variables (e.g., \textit{measurement variables} of system internal states) during model construction. Faults typically manifest in system-dependent variables, while system-independent variables, which contain information on operating conditions, may remain unaffected by faults but significantly influence system-dependent variables. Incorporating the context of operating conditions into node dynamics extraction enables the model to more accurately separate actual faults from novel operating conditions.
}
\revtextrewrite{
To summarize, our key contributions are as follows:
\begin{itemize}
\item \textbf{Dynamic edge construction for MTS graph inference}: The proposed DyEdgeGAT algorithm dynamically constructs edges between time series signals, enabling the model to capture the evolution of pairwise relationships.
\item \textbf{Operating condition-aware node dynamics modeling}: System-independent variables in the node dynamics extraction and reconstruction are modeled in different ways within DyEdgeGAT, enabling the distinction between faults and novel operating conditions.
\item \textbf{Temporal topology-informed anomaly scoring}: The proposed anomaly score incorporates temporal topology to account for the diverse strengths in sensor dynamics within IIoT systems with heterogeneous signals.
\item \textbf{Comprehensive performance evaluation}: We evaluated the proposed DyEdgeGAT algorithm on synthetic and real-world datasets across varying fault severities, multiple fault types, and novel operating conditions and compared it to a wide range of algorithms.
\end{itemize}
}

\revtextiot{
The remainder of this paper is organized as follows:
Sec.~\ref{sec:rw} reviews related work in unsupervised fault and anomaly detection for time series data, graph learning from multivariate time series, and GNN-based fault and anomaly detection. Sec.~\ref{sec:method} elaborates on DyEdgeGAT's core components. Sec.~\ref{sec:case_study} introduces the case studies along with data statistics. Sec.~\ref{sec:exp_design} outlines the experimental design, including baseline methods and the evaluation metrics. Sec.~\ref{sec:results} presents and discusses our results. Finally, Sec.~\ref{sec:conclusion} provides conclusions and suggests future research directions.}

%% file: tabs/overview.tex
\begin{table}[tb]
\centering
{\color{revcolor} 
\caption{{Comparative Analysis of Approaches for Relationship Modelling in Multivariate Time Series}}
\label{tab:intro_overview}
\begin{tabular}{lccc}
\hline
\multirow{2}{*}{\textbf{{Type of approaches}}} & \textbf{{Spatial-}} & \textbf{{Temporal}} & \textbf{{Evolving}} \\
                                      & \textbf{{dependencies}} & \textbf{{dynamics}} & \textbf{{relationships}} \\
\hline
{Relationship-focused}   & {$\checkmark$}            & {-}                & {-}            \\
{Dynamics-focused}       & {-}                   & {$\checkmark$}         & {-}            \\
{Graph-based}            & {$\checkmark$}            & {$\checkmark$}         & {-}            \\
{DyEdgeGAT (Proposed)}   & {$\checkmark$}            & {$\checkmark$}         & {$\checkmark$}     \\
\hline
\end{tabular}
}
\end{table}

%% file: chaps/02-rw/02-related_work.tex
\section{Related Work}
\label{sec:rw}
\revtextiot{
This section reviews relevant areas of our proposed method from both application and methodological perspectives.
We begin with a discussion of general unsupervised fault and anomaly detection methods for time series (Sec.~\ref{sec:rw_fd}), chosen for their foundational relevance in understanding and identifying irregular patterns and behaviors in time-series data.
Next, we review methods for multivariate time series graph construction and temporal graph representation learning (Sec.~\ref{sec:graph_tgnn}). This area is selected due to its importance in effectively representing complex relationships in time series data, a core aspect of our approach. 
Lastly, we examine the application of Graph Neural Networks (GNNs) in detecting faults and anomalies (Sec.~\ref{sec:rw_gnn_ad}), 
highlighting the recent advances in GNN techniques and their relevance to complex system analysis.
}

\input{chaps/02-rw/02-1-rw-ad}

\input{chaps/02-rw/02-2-graph_tgnn}
\input{chaps/02-rw/02-4-gnn_ad}

%% file: chaps/02-rw/02-1-rw-ad.tex
\subsection{Fault and Anomaly Detection for Time Series Data}
\label{sec:rw_fd}
Given the rarity of faults, this section focuses on reviewing unsupervised Fault Detection (FD) techniques. Additionally, Anomaly Detection (AD) methods are also reviewed, due to their methodological similarity to unsupervised FD.

\textbf{Unsupervised fault detection}. Traditional unsupervised FD has focused on identifying patterns in data that deviate from its normal condition and shifts in data interdependencies. Classical machine learning methods, such as one-class Support Vector Machines (SVMs), have been commonly applied in FD, effectively identifying outliers by enclosing all positive instances within a hyper-sphere~\cite{yin2014ocsvm}. Autoencoders (AEs) are popular in unsupervised FD, facilitating fault detection through monitoring the deviation in the reconstruction errors~\cite{plakias2022novel}\cite{guo2020unsupervised}. Generative Adversarial Networks (GANs) have also been utilized for FD, with the generator creating positive samples and the discriminator differentiating between normal and abnormal instances~\cite{akcay2019ganomaly}\cite{plakias2019exploiting}.

\textbf{Anomaly detection.} In the IIoT context, AD methods focus on modeling system dynamics and identifying deviations from these as anomalies. Approaches such as Recurrent Neural Networks (RNNs)~\cite{li2020anomaly}, Long Short-Term Memory (LSTM) networks\cite{wu2019lstm}, and Convolutional Neural Networks (CNNs)~\cite{garcia2022temporal} are commonly employed, using prediction or reconstruction errors as anomaly indicators. These anomalies typically manifest as point anomalies, often due to sensor faults, or context anomalies, which often arise from operational misconfigurations or changes in environmental conditions~\cite{cook2019anomaly}\cite{chatterjee2022iot}. Additionally, self-supervised methods like Masked Anomaly Detection (MAD) have been developed, enabled by input sequence masking and estimation~\cite{fu2022mad}.

\textbf{Novel operating conditions.} A common challenge in FD and AD is their inability to distinguish between novel operating conditions and faults, which arises from a limited representation of all normal operating conditions in the training data. To address this, Guo \textit{et al.} proposed to combine clustering with expert input to better account for unknown operating modes. Differently, Michau \textit{et al.}~\cite{michau2019unsupervised} proposed to utilize unsupervised feature alignment to extract features under varying conditions and integrate them for fault detection. Expanding upon this approach, \revtextiot{Rombach \textit{et al.}~\cite{rombach2021contrastive} enhanced the feature representation learning with contrastive learning using triplet loss to achieve invariance to novel operating conditions.}

Although state-of-the-art FD and AD methods have demonstrated a good performance in modeling interdependencies and dynamics within the data, they fall short in modeling the evolution of relationships in the system. This deficiency affects their ability to detect faults characterized by relationship shifts, especially incipient faults. 
In addition, to account for novel conditions, current approaches either require labeled operating conditions in the training phase or require explicit modeling to extract operating condition invariant features. Addressing dynamic modeling that can implicitly model operating conditions remains a challenge.

%% file: chaps/02-rw/02-2-graph_tgnn.tex
\subsection{Graph Neural Networks for Multivariate Time Series}
\label{sec:graph_tgnn}
\revtext{
Graph-based methods are effective tools for modeling complex spatial-temporal dynamics in MTS. Two crucial aspects underlie this process: \textit{graph inference}, involving the construction of graph structures from MTS, and the application of \textit{graph neural networks} for subsequent analysis such as forecasting, imputation or anomaly detection~\cite{jin2023survey}.

\textbf{Graph inference} is the initial step to transform MTS data into temporal graphs, also referred to as spatial-temporal or dynamic graphs (with no distinction among these terms in our context).
In practice, two types of strategies are utilized to construct temporal graphs from MTS data: \textit{heuristics} or \textit{learned} from data~\cite{jin2023survey}. Heuristic-based methods extract graph structures from data based on heuristics such as spatial connectivity~\cite{li2017diffusion} and pairwise similarity~\cite{chen2023multivariate}.
Learning-based methods directly learn the graph structure from the data in an end-to-end fashion. These methods commonly utilize embedding-based~\cite{wu2020connecting}, attention-based~\cite{zhao2020mtadgat}, and sampling-based methods~\cite{zhang22grelen}. These learning-based approaches enable the discovery of more complex and potentially more informative graph structures compared to heuristic-based methods \cite{zugner2021study}.

\textbf{Graph neutral networks} are employed to process the spatial-temporal dynamics captured in the inferred temporal graphs.
We follow the definition by Gao et al.~\cite{gao2022equivalence}, distinguishing between two primary paradigms: \textit{time-and-graph}, where graph representations derived from GNNs are integrated with sequence models like RNNs to jointly capture the temporal dynamics of node attributes; and \textit{time-then-graph}, where sequences that describe node and edge dynamics are first modeled and then incorporated as attributes in a static aggregated graph representation.}
\revtextrewrite{
Existing GNN methods for MTS mainly follow the \textit{time-and-graph} approach, dynamically constructing a static graph for each input sequence. For instance, Diffusion Convolutional Recurrent Neural Network (DCRNN)~\cite{li2017diffusion} applies SpectralGCN and GRU to a predefined static graph for traffic forecasting.  Extending this, Graph for Time Series (GTS)~\cite{shang2021gts} builds upon DCRNN, employing it to a jointly learned probabilistic global graph. 
In the \textit{graph-then-time} category, a variation of \textit{time-and-graph}, Anomaly Detection via Dynamic Graph Forecasting (DyGraphAD)~\cite{chen2023multivariate} is worth noting. Specifically, it generates a series of dynamic correlation graphs from MTS using dynamic time warping and processes these graphs to create a sequence of latent graph representations for forecasting.
Conversely, \textit{time-then-graph} has mainly been applied to networks with predefined graph structures such as evolving social networks or traffic systems, where simultaneous graph inference is not required. Notable examples of the \textit{time-then-graph} frameworks include Temporal Graph Attention Networks (TGAT)~\cite{xu2020tgat} and Temporal Graph Network (TGN)~\cite{rossi2020tgn}.}

\revtext{
Given the complex nature of MTS data in the IIoT context, more expressive modeling techniques are crucial to improve fault and anomaly detection performance. Gao \textit{et al.}~\cite{gao2022equivalence} have demonstrated the superior expressiveness of the \textit{time-then-graph} approach for MTS. However, applying it to IIoT settings introduces a research gap: it requires the extraction of edge dynamics to infer the temporal graph, which is different from the static graph inference common in the existing literature. 
While some \textit{time-and-graph} approaches do construct dynamic graphs from windowed MTS data, they typically produce a single static graph per input, failing to track evolving edge dynamics and graph structures. 
Conversely, DyGraphAD constructs dynamic graphs based on the correlation of MTS but follows a \textit{graph-then-time} approach, emphasizing graph embedding changes over the evolution of pairwise relationships which makes them less applicable to detect faults characterized by relationship changes.
}

%% file: chaps/02-rw/02-4-gnn_ad.tex
\subsection{GNN-based Time Series Fault and Anomaly Detection}
\label{sec:rw_gnn_ad}
\revtextrewrite{
Recent works have explored GNN for MTS anomaly detection. For instance,  Deng \textit{et al.}~\cite{deng2022graph} introduced the Spatio-temporal Graph Convolutional Adversarial Network (STGAN)~\cite{deng2022graph}, employing a spatiotemporal generator and discriminator to address the challenge of traffic anomalies with varying criteria across locations and time, thereby enhancing early detection capabilities.
Following this trend, Constant-Curvature Riemannian Manifolds Change Detection Test (CCM-CDT)~\cite{zambon2018concept} utilizes an adversarially trained graph autoencoder. This autoencoder generates latent space points on Riemannian manifolds where statistical tests are performed to identify stationarity changes in graph streams. 
Building upon the concept of special analysis, Graph Wavelet Variational Autoencoder (GWVAE)~\cite{li2023novel} utilizes spectral graph wavelet transform which can realize multiscale feature extraction for FD.
To better capture the complex spatial-temporal dependencies in MTS, Time-series Anomaly Detection via Graph Attention Network (MTATGAT)~\cite{zhao2020mtadgat} leverages feature-oriented and time-oriented graph attention mechanisms in a joint reconstruction and forecast discrepancy framework to simultaneously capture spatial and temporal dependencies for more accurate AD.
Similarly, Graph Learning with Transformer for Anomaly Detection (GTA)~\cite{chen2021learning} is a transformer-based forecasting-based approach for cyber attack detection in IIoT systems, leveraging a new graph convolution called influence propagation to simulate the information flow among the sensors.
To address the heterogeneity in IIoT sensors, GDN~\cite{deng2021gdn} utilized sensor embeddings for graph construction and prediction-based detection.
Extending the learned spatial correlation from GDN, Correlation-aware Spatial-Temporal Graph Learning (CST-GL)\cite{zheng2023correlation} further exploited multi-hop graph convolution as well as dilated Temporal Convolutional Network (TCN) to capture long-range dependence over space and time.
Another notable example of GNN-based AD methods is Graph Representation Learning for Anomaly Detection (GRELEN)~\cite{zhang22grelen}, which was the first to propose AD based on graph relation discrepancy. It utilized VAE to learn probabilistic graph relations.
}

\revtext{
While several GNN-based approaches have demonstrated effectiveness in AD within Multivariate Time Series, they face two main limitations that need to be addressed. Firstly, these methods primarily employ \textit{time-and-graph} representations and focus on node dynamics via forecasting or reconstruction. These approaches have been effective in detecting anomalies arising from deviated system dynamics, particularly in scenarios where single sensor behaviors (point anomalies) and temporal patterns (context anomalies) are key indicators.
Secondly, and more crucially, these methods exhibit limitations in capturing early-stage faults characterized by relationship shifts. This limitation stems from discrete static graphs generated for each input sequence, which neglects the temporal evolution of the graph structure and potentially misses subtle relationship shifts. To the best of our knowledge, no existing GNN methods effectively address these relationship shifts within MTS.
}

%% file: chaps/03-fw/03-framework.tex
\section{Proposed Framework}
\label{sec:method}

\input{chaps/03-fw/03-1-overview}

\input{chaps/03-fw/03-2-dynamic-edge}
\input{chaps/03-fw/03-3-oc}

\input{chaps/03-fw/03-4-gnn}
\input{chaps/03-fw/03-5-recon}
\input{chaps/03-fw/03-6-loss}
\input{chaps/03-fw/03-7-ad_score}

%% file: chaps/03-fw/03-1-overview.tex
\revtextrewrite{
In this paper, we utilize bold uppercase letters (e.g., $\mathbf{X}$), bold lowercase letters (e.g., $\mathbf{x}$), and calligraphic letters (e.g., $\mathcal{V}$) to denote matrices, vectors, and sets, respectively.

\subsection{Problem Statement}
In an IIoT sensor network, $N$ system-dependent measurement signals at any time $t$ are represented as $\mathbf{x}^{t} = [x_{1}^{t}, \cdots, x_{N}^{t}] \in \R^{N}$, where $x_{i}^{t}$ represents the $i^{th}$ sensor's time series at time $t$. 
By employing a sliding window approach of length $W$, we construct samples as:
$\mathbf{X}^{t_w:t} = [\mathbf{x}^{t_w}, \cdots, \mathbf{x}^{t-1}, \mathbf{x}^{t}] \in \R^{N \times W}$, where $t_w = t - W + 1 > 0$. 
Similarly, system independent variables (control variables and external variables) are denoted $\mathbf{U}^{t_w:t} \in \R^{N_u \times W}$.
The proposed method, DyEdgeGAT, employs a reconstruction model $f_\theta$ for unsupervised MTS fault and anomaly detection.
The model generates an output $\mathbf{\hat{X}}^{t_w:t} = f_\theta(\mathbf{X}^{t_w:t}, \mathbf{U}^{t_w:t})$ for each $N_{train}$ healthy samples, aiming to minimize the reconstruction discrepancy $||\mathbf{\hat{X}}^{t_w:t} - \mathbf{X}^{t_w:t}||$.
At test time, the model outputs $\mathbf{y} \in \R^{N_{test}}$ for all $N_{test}$ samples based on the reconstruction discrepancy, where each element $\mathbf{y}_i \in \{0, 1\}$ indicates faults.
}

\subsection{Framework Overview}
\begin{figure*}[tbhp]
\centering
    \centering
    \includegraphics[width=\linewidth]{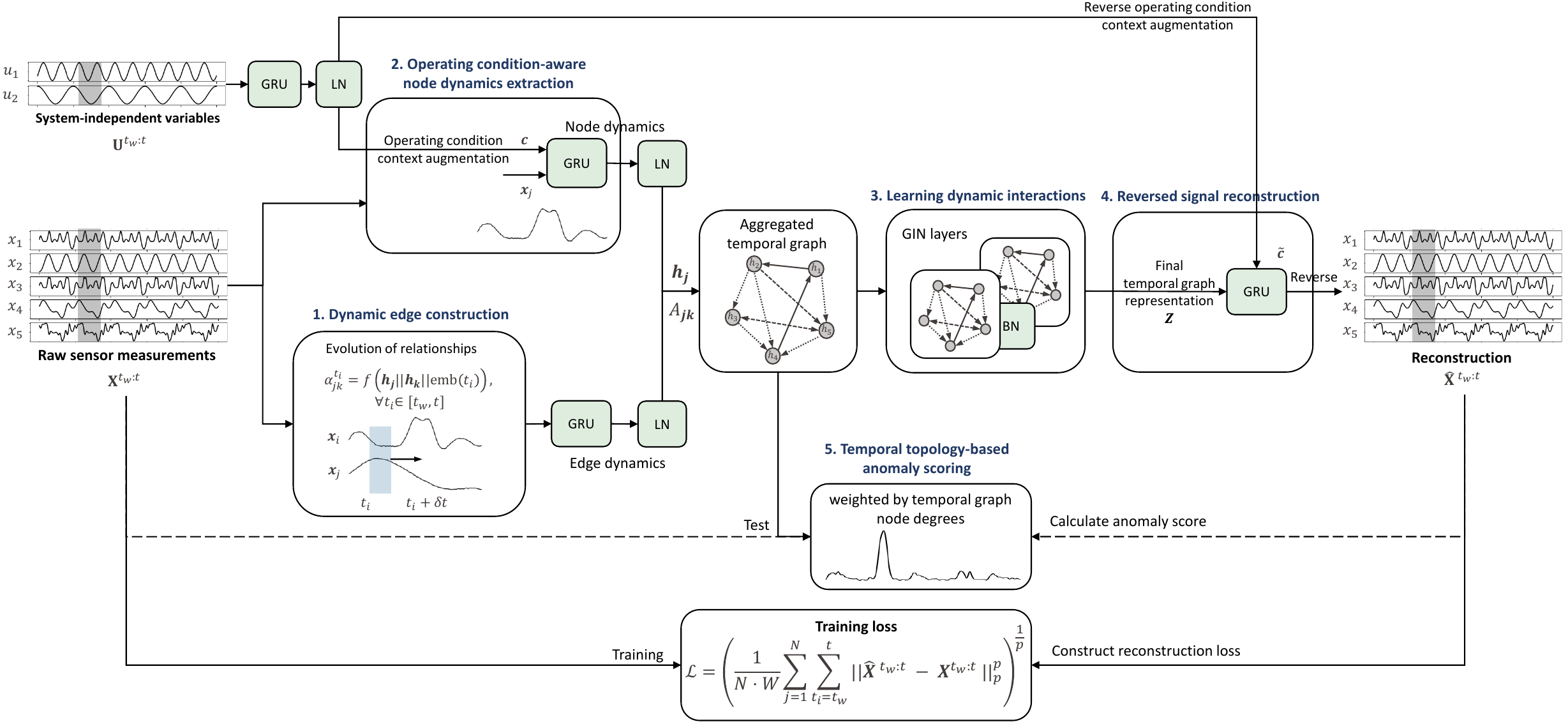}
    \caption{Overview of the Dynamic Edge via Graph Attention (\textbf{DyEdgeGAT}). 
    Starting from raw sensor measurements $\mathbf{X}^{t_w:t}$ and system-independent variables $\mathbf{U}^{t_w:t}$, the process involves: \revtextiot{\textbf{(1) Dynamic edge construction}}, where the model infers and tracks evolving interdependencies between time series. \revtextiot{\textbf{(2) Operating condition aware node dynamics extraction}}, augmented by operating condition context via GRU and Layer Normalization (LN) modules. \revtextiot{\textbf{(3) Dynamic interaction learning}}, with two Graph Isomorphism Network (GIN) layers and a Batch Normalization (BN) layer in between. \revtextiot{\textbf{ (4) Reverse signal reconstruction} augments operating condition context and reconstructs} the original sensor signals in the reversed order. \revtextiot{\textbf{(5) Temporal topology-informed anomaly scoring}}, leveraging the learned temporal graph structure to balance different strengths of dynamics in the heterogeneous signals. In the training phase, the model minimizes reconstruction loss using normal data. During the testing phase, the model employs reconstruction discrepancies, adjusted by interaction strengths among sensor nodes for anomaly scoring.}    
    \label{fig:dyedgegat_overview}
\end{figure*}

\revtextrewrite{
We propose \textit{Dynamic Edge via Graph Attention} (\textbf{DyEdgeGAT}) for fault detection to overcome the limitations of existing GNN-based methods. 
To address the complexity of temporal and spatial dynamics in IIoT systems, DyEdgeGAT employs the \textit{time-then-graph} framework based on the aggregated temporal graph representation. 
The \textit{time-then-graph} framework in DyEdgeGAT, as the name suggests, sequentially extracts temporal patterns using a sequence model and captures spatial-temporal relationships using GNNs. These two steps are elaborated in detail below.
Unlike traditional models that treat edge relationships as static and focus mainly on node dynamics, DyEdgeGAT employs dynamic edge construction to adaptively capture evolving temporal interdependencies. This enables the model to recognize relationship shifts at early fault stages.
Another key innovation of DyEdgeGAT is its integration of operating condition context into node dynamics extraction. This integration improves the model's robustness under varying operating conditions as well as helps the model distinguish between faults and novel operating conditions. 
An overview of the DyEdgeGAT framework can be found in Fig.~\ref{fig:dyedgegat_overview}.
DyEdgeGAT consists of five core components, each elaborated in subsequent sections:
\begin{enumerate}
\item \textbf{Dynamic edge construction} enables a novel graph inference scheme for MTS that dynamically constructs edges to represent and track the evolving relationships between time series at individual time steps (Sec.~\ref{sec:method_graph_constr}).
\item \textbf{Operating condition-aware node dynamics extraction} innovatively incorporates operating condition contexts into node dynamics, mitigating false alarms due to novel operating conditions (Sec.~\ref{sec:method_oc_node}).
\item \textbf{Dynamic interaction modeling} utilizes GNNs on the inferred aggregated temporal graph, integrating both node and edge dynamics to capture evolving interactions (Sec.~\ref{sec:method_graph_repr}).
\item \textbf{Reversed signal reconstruction} reconstructs sensor signals, enhanced with reversed operating condition contexts for robust reconstruction (Sec.~\ref{sec:method_recon}).
\item \textbf{Temporal topology-informed anomaly scoring} leverages the temporal graph topology to normalize the anomaly score, taking into account different intensities of dynamics in the heterogeneous signals (Sec.~\ref{sec:method_ad_score}).
\end{enumerate}
}

%% file: chaps/03-fw/03-2-dynamic-edge.tex
\subsection{Dynamic Edge Construction with Attention Mechanism}
\label{sec:method_graph_constr}
\revtext{
The dynamic edge construction module focuses on inferring and representing MTS as temporal graphs. We aim to leverage the equivalence between two common temporal graph representations to effectively represent the relationship changes in MTS.
Following the terminology of Gao \textit{et al.}~\cite{gao2022equivalence}, the two equivalent temporal graph representations are defined as follows. 
The first, \textbf{discrete-time dynamic graph}, represents MTS data as a sequence of attributed graphs over discrete time steps: $\mathcal{G} = \{\mathcal{G}^{t_w}, \ldots, \mathcal{G}^t\}$, where each graph $\mathcal{G}^{t_i}(\mathbf{x}^{t_i}, A^{t_i})$  at time instance $t_i$ is defined by its feature matrix $\mathbf{x}^{t_i} \in \R^{N}$ and adjacency matrix $\mathbf{A}^{t_i}\in \R^{N \times N}$. 
The second, \textbf{aggregated temporal graph}, models MTS as a static graph aggregating node and edge attributes over time: $\mathcal{G} = (\mathbf{X}^{t_w:t}, \mathbf{A}^{t_w: t})$, with $\mathbf{X}^{t_w:t} \in \R^{N\times W}$ and $\mathbf{A}^{t_w:t} \in \R^{N\times N \times W}$ representing aggregated node and edge attributes, respectively.
By leveraging the equivalence of both graph representations, we construct discrete-time graph snapshots within sliding windows $[t_i, t_i +\delta t]$ for each input sequence interval  $[t_w, t]$ using an edge weight matrix $\mathbf{A}^{t_i}$, and then integrate these into an aggregated graph.
}
\revtext{Prior to graph inference, each input node signal $\mathbf{x}_j \in \mathbf{X}$, is pre-processed using a 1D Convolutional Neural Network (1DCNN). The 1DCNN preserves the dimensionality of the input sequence, outputting denoised feature vectors $\mathbf{h}_j = \text{1DCNN}(\mathbf{x}_j)\in \mathbb{R}^W$ for each node $j$, preserving the sequence length while reducing noise, enhancing the robustness of the edge construction process.}

\revtextrewrite{
Edge weights are inferred using a GATv2-based attention mechanism~\cite{brody2022gatv2}, generating a set of attention coefficients $\{\alpha_{jk}^{t_w}, \dots, \alpha_{jk}^{t}\}$ for each time window $[t_w, t]$. To infer the attention score per timestamp, we 
 first introduce the edge significance scoring function $e: \R^W\times \R^W\times \R^d \rightarrow \R$ to compute the relevance of node $k$'s features to node $j$ at time step $t_i$, defined as:
\begin{equation}
    e_{jk}^{t_i} = \mathbf{a}^T \text{LeakyReLU}\left(\mathbf{W} \cdot \left[\mathbf{h}_{j} \parallel \mathbf{h}_{k} \parallel \text{emb}(t_i)\right]\right), \label{eq:score_fun}
\end{equation}
where $\mathbf{a}\in \R^{2d'}$ and $\mathbf{W}\in \R^{d' \times (2W+d)}$ are learnable parameters, and $\parallel$ denotes vector concatenation.
Additionally, to ensure that the edge scores are comparable across different nodes, we normalize them across all neighbors $k\in \mathcal{N}_j$ using the softmax function to obtain the attention coefficients $\alpha_{jk}^{t_i}$:
\begin{equation}
    \alpha_{jk}^{t_i} = \text{softmax}_j \left( e_{jk}^{t_i} \right) = \frac{\exp \left(e_{jk}^{t_i} \right)}{\sum_{k^{\prime} \in \mathcal{N}_j} \exp \left(e_{jk^{\prime}}^{t_i}\right)}.
\end{equation}
Here, $\mathcal{N}_j$ denotes the set of neighbors for node $j$. The attention coefficients $\alpha_{jk}^{t_i}$ are then used as the edge weights $A_{jk}^{t_i}$ for time step $t_i$, thereby, capturing the importance of each edge in the graph.
In contrast to traditional static attention-based graph inference, as used in MTADGAT~\cite{zhao2020mtadgat}, our proposed approach in Eq.~\ref{eq:score_fun} integrates a novel term $\text{emb}(t_i)$, which transforms a real-value timestamp into a vector, providing a context of the temporal position. 
The formulation of $\text{emb}(t_i)$ aligns with the temporal encoding used in TGN \cite{rossi2020tgn}, which is defined as:
\begin{equation}
    \text{emb}(t_i) = \left[ \cos(\omega_0 t_i), \cos(\omega_1 t_i), \ldots, \cos(\omega_d t_i) \right], \label{eq:time_enc}
\end{equation}
with frequencies $\omega_i = \frac{1}{10^{i/d}}$, where $i$ ranges from $1$ to time embedding dimension $d$. 
Although both our method and TGN employ the same cosine-based temporal encoding, the intuition behind them differs. Specifically, We encode absolute timestamps to track attention evolution over time while TGN encodes relative timestamps to learn time-invariant features.
}
\revtext{
Finally, we apply a GRU-based mechanism to model the temporal evolution of edge weights (attention coefficients $\mathbf{\alpha}_{jk}^{t_i}$):
\begin{equation}
    \mathbf{h}_{jk}^{t_i} = \text{ReLU}\left(\text{GRU-Cell}(\mathbf{\alpha}_{jk}^{t_i}, {\mathbf{h}}_{jk}^{t_i-1})\right), \forall t_i \in [t_w, t].
    \label{eq:edge_gru}
\end{equation}
Here, \({\mathbf{h}}_{jk}^{t_i} \in \mathbb{R}\) is the hidden state for edge $(j, k)$ at time \(t_i\), with the initial state ${\mathbf{h}}_{jk}^{t_w-1}$ set to $\mathbf{0}$. The final state ${\mathbf{h}}_{jk}^{t}$ representing the encoded edge dynamics is then used in Eq.~\ref{eq:gnn} as the edge weight $\mathbf{A}\in \mathbb{R}^{N\times N}$ for the static temporal graph.
}

%% file: chaps/03-fw/03-3-oc.tex
\subsection{Operating-Condition-Aware Node Dynamics Extraction}
\label{sec:method_oc_node}
\revtext{
In our approach, we emphasize the importance of incorporating operating condition contexts into the process of node dynamics extraction, a critical aspect for enhancing fault detection and ensuring robustness against novel operating conditions. 
Traditional AD algorithms often detect new operating conditions as anomalies due to significant deviations from the known dynamics of the training dataset. 
By integrating these operational state contexts, our method aims to address this limitation.
Node dynamics extraction in our framework focuses on extracting the dynamics of each individual sensor.
Our approach distinctively distinguishes between measurements that are dependent on the system $\mathbf{X}$ from system-independent variables ($\mathbf{U}$), such as control inputs and external factors in this process.
While $\mathbf{X}$ reflects the system's current state through sensor measurements, $\mathbf{U}$ presents control inputs and external factors. Unlike measurements, $\mathbf{U}$ representing the system's operating state, expresses a strong influence on the system dynamics but remains largely unaffected by system faults.}

\revtext{
We propose to extract operational state context with a Gated Recurrent Unit (GRU) network, which processes the sequence of \(\mathbf{U}^{t_w:t}\). The GRU updates its cell state at each time step to capture temporal dynamics:
\begin{equation}
    \mathbf{h}_c^{t_i} = \text{ReLU}\left(\text{GRU-Cell}(\mathbf{U}^{t_i}, \mathbf{h}_c^{t_i-1})\right), \forall t_i \in [t_w, t],
    \label{eq:control_gru}
\end{equation}
where \(\mathbf{h}_c^{t_i} \in \mathbb{R}^{d_h}\) represents the hidden state at time \(t_i\) with \(d_h\) being the hidden state dimensionality. 
The initial hidden state is set to zero, $\mathbf{h}_c^{t_w-1} = \mathbf{0}$. The final hidden state $\mathbf{h}_c^t$ encodes the dynamic operational state context and is then used to initialize the node encoder for each node $j$:
\begin{equation}
    \mathbf{h}_j^{t_i} = \text{ReLU}\left(\text{GRU-Cell}(\mathbf{x}_j^{t_i}, \mathbf{h}_j^{t_i-1})\right), \forall t_i \in [t_w, t].
    \label{eq:node_gru}
\end{equation}
Here, \(\mathbf{h}_j^{t_i} \in \mathbb{R}^{d_h}\) is the hidden state for node \(j\) at time \(t_i\), with the initial state \(\mathbf{h}_j^{t_w-1}\)  set to $\mathbf{h}_c^t$. The final state $\mathbf{h}_j^t$ represents the encoded dynamics of node $j$ up to time $t$ with the context of operating conditions. The final states of all nodes form the hidden node presentation $\mathbf{H}\in \mathbb{R}^{N\times d_h}$ in Eq.~\ref{eq:gnn}.
}
Note that the GRU in Eq.~\ref{eq:control_gru} is a multivariate GRU, processing all control variables, whereas the GRU in Eq.~\ref{eq:node_gru} is a univariate GRU with shared weights across all nodes.

%% file: chaps/03-fw/03-4-gnn.tex
\subsection{Dynamic Interaction Modeling}
\label{sec:method_graph_repr}
\revtext{
Dynamic interaction modeling focuses on modeling dynamic system interactions using GNNs.
To integrate the dynamics in the learning architecture, we propose to employ GNNs on a static weighted aggregated temporal graph $\mathcal{G}(\mathbf{H}, \mathbf{A})$, that reflects system dynamic states across various operating conditions. 
This graph aggregates a hidden node representation containing operating condition-aware node dynamics (Eq.~\ref{eq:node_gru}) as node attributes, denoted by $\mathbf{H} \in \mathbb{R}^{N \times N_{d_h}}$, and a hidden edge representation encapsulating edge dynamics (Eq.\ref{eq:edge_gru}) as edge weights, represented by $\mathbf{A} \in \mathbb{R}^{N \times N}$.
The interaction learning process applies multiple GNN layers to extract hidden node representations $\mathbf{z}_j \in \mathbb{R}^{d_z}$:
\begin{align}
\mathbf{Z} &= \text{GNN}^L(\mathbf{H}, {\mathbf{A}}). \label{eq:gnn}
\end{align}
Here, $\mathbf{Z} \in \mathbb{R}^{N\times d_z}$ is the final node representation after $L$ layers of GNN processing, where $d_z$ is the dimensionality of the output space, and $\mathbf{z}^0$ is initialized with  $\mathbf{H}$.
}
\revtext{
The GNN layer in Eq. \ref{eq:gnn} follows the Message Passing Neural Network (MPNN) schema~\cite{gilmer2017neural}.  At the $l$-th GNN layer, message passing and update steps occur as follows:
\begin{align}
    \mathbf{m}^{(l)}_j &= \sum_{j \in \mathcal{N}_j} \text{MSG}^{(l)} (\mathbf{z}^{(l-1)}_j, \mathbf{z}^{(l-1)}_k, {A}_{jk}), \\
    \mathbf{z}^{(l)}_j &= \text{UPDATE}^{(l)} (\mathbf{z}^{(l-1)}_j, \mathbf{m}^{(l)}_j).
\end{align}
Here, $\mathcal{N}_j$ denotes the neighborhood of node $j$, $\mathbf{m}^{(l)}_i$ is the message accumulated at node $j$ in layer $l$, and ${A}_{jk}$ represents the edge weights between nodes $j$ and $k$ of the aggregated temporal graph. The functions $\text{MSG}$ and $\text{UPDATE}$ are learnable transformations specific to the GNN architecture, which aggregate information from the neighboring nodes and update the node representation, respectively.
Specifically, we utilize Graph Isomorphism Network (GIN)~\cite{hu2019gine} as the GNN module, which allows the $\text{MSG}$ function to incorporate edge weights.
The $\text{UPDATE}$ function is a two-layer MLP (Multi-Layer Perceptron) with Batch Normalization (BN)
Node $j$'s representation at layer $l$, $\mathbf{z}^{(l)}_j$, updates as:
\begin{equation}
    \mathbf{z}^{(l)}_j = \text{MLP}^{(l)}\left( (1 + \epsilon^{(l)}) \cdot \mathbf{z}^{(l-1)}_j + \sum_{k \in \mathcal{N}(j)} {A}_{jk} \cdot \mathbf{z}^{(l-1)}_k \right). \label{eq:gnn_z}
\end{equation}
Here, $\epsilon^{(l)}$ is a learnable parameter at layer $l$ that allows the model to weigh self-connections. BN is employed between GIN layers for normalization.
}

%% file: chaps/03-fw/03-5-recon.tex
\subsection{Reversed Signal Reconstruction}
\label{sec:method_recon}
\revtextrewrite{
DyEdgeGAT adopts an innovative approach for fault detection by reconstructing the measurement variables ($\mathbf{X}$) in a reversed order, inspired by a technique used in the Seq2Seq model for neural machine translation~\cite{cho2014learning}. This novel reversal technique helps to effectively align temporally distant but causally relevant features, thereby facilitating learning in longer sequences where gradient vanishing during weight backpropagation poses a challenge.
The reconstruction begins with the final graph representation (Eq.~\ref{eq:gnn_z}), which encodes system dynamic states across operating conditions.
For reconstructing the original sequence of measurement variables $\mathbf{x}_j^{t_w:t}$, a GRU network is repurposed.  
Intially, $\mathbf{z}_j$ undergoes a linear transformation $\mathbf{z}_j^{\prime} = \mathbf{W}_z \mathbf{z}_j + \mathbf{b}_z$, yielding $\mathbf{z}_j^{\prime} \in \mathbb{R}^{d_h}$. 
The GRU then processes the transformed node features $\mathbf{z}_j^{\prime}$ in reverse order:
\begin{equation}
\overleftarrow{\mathbf{h}}_{o, j}^{t_i} = \text{ReLU}\left(\text{GRU-Cell}(\mathbf{z}_j^{\prime}, \overleftarrow{\mathbf{h}}_{o, j}^{t_{i}-1})\right), \forall t_i \in [t, t_w].
\end{equation}
Initialization of the GRU cells uses $\overleftarrow{\mathbf{h}}_{o, j}^{t-1} = \overleftarrow{\mathbf{h}}_c^{t_w}$, the final hidden state from the reversed control variable sequence $\mathbf{U}^{t_w:t}$ (Eq.~\ref{eq:control_gru}) in a similar fashion. The predicted sequence for node $j$ is then reconstructed at each timestep $t_i$:
Similarly, we inverse the operating condition context as well, and use $\overleftarrow{\mathbf{h}}_{o, j}^{t-1} = \overleftarrow{\mathbf{h}}_c^{t_w}$ to initialize the GRU cells, which is the last hidden state of the GRU for the reversed control variable sequence $\mathbf{U}^{t_w:t}$ in Eq.~\ref{eq:control_gru}.
Finally, we reconstruct the predicted sequence for each node $j$ through a linear output layer for each timestep $t_i$:
\begin{equation}
{\hat{x}}_j^{t_i} = \mathbf{W}_o \overleftarrow{\mathbf{h}}_{o, j}^{t_i} + \mathbf{b}_o, \forall t_i \in [t_w, t]
\end{equation}
where $\mathbf{W}_o \in \mathbb{R}^{1 \times d_h}$ and $\mathbf{b}_o \in \mathbb{R}$ form the linear output layer. 
}


%% file: chaps/03-fw/03-6-loss.tex
\subsection{Training Objective}
\revtext{
The objective of the training is to minimize the discrepancy between the reconstructed sequence \({\hat{x}}_j^{t_i}\) and the true sequence \({x}_j^{t_i}\) across all sensors and the entire sliding window. The \( p \)-norm loss function is defined as:
\begin{equation}
\mathcal{L}=  \left(\frac{1}{N \cdot W} \sum_{j=1}^{N} \sum_{t_i=t_w}^{t} \left\| \hat{x}_j^{t_i} - x_j^{t_i} \right\|_p^p \right)^{1/p}, p \geq 1
\end{equation}
where $N$ is the number of nodes, $W$ is the sliding window length, and $p$ is the norm degree. The parameters of the reconstruction model \( f_\theta \) are optimized to minimize this loss across all training samples \( N_{train} \).
}



%% file: chaps/03-fw/03-7-ad_score.tex
\subsection{Temporal Topology-Based Anomaly Score Design}
\label{sec:method_ad_score}
\revtext{
In heterogeneous IIoT environments, sensors exhibit diverse dynamic behaviors, which can impact signal reconstruction quality and, consequently, have a negative impact on fault and anomaly detection. Anomaly scoring functions derived directly from reconstruction errors are often biased toward sensors with more significant dynamics due to their larger error magnitudes.
To address this, we propose a temporal topology-based anomaly score, where ``temporal topology'' refers to the interaction strengths among sensor nodes over time, encoded in the aggregated temporal graph. This structure not only reflects the strength of these interactions but also their evolution in time. Changes in interaction strength with other signals are indicative of potential faults, offering an effective method for identifying subtle faults.}
\revtextrewrite{
Building on the concept of strength in dynamic interaction, our proposed methodology normalizes the reconstruction error of each sensor signal sequence $\mathbf{x}_j \in \mathbf{X}^{t_w:t}$ by its corresponding node's degrees in the graph, reflecting its interaction strength with other signals:
\begin{equation}
\mathbf{r}_j = \frac{1}{d_j} |\hat{\mathbf{x}}_i-\mathbf{x}_j|
\end{equation}
Here, $d_j$ represents the sum of the weighted in-degree and out-degree of node $j$, capturing the intensity of its interactions across the graph.
The final anomaly score, $s$, is calculated by averaging these topology-normalized reconstruction errors across all sensors and the entire sequence length, given by:
\begin{equation}
s = \frac{1}{N}\frac{1}{W}\sum_{j=1}^N\sum_{t_i=t_w}^t r_j^{t_i}
\end{equation}
Anomalies are then identified based on $s$, using a threshold determined empirically or via statistical analysis of the healthy validation set.
}

%% file: chaps/04-doe/04-1-casestudies.tex
\section{Case Studies}
\label{sec:case_study}
\revtextiot{
We evaluate DyEdgeGAT based on two case studies to assess its efficacy and robustness in fault detection within complex systems. 
The first case study employs a synthetic dataset, developed due to the need to fully control faults characterized by relationship shifts. It provides a unique opportunity to access the ground truth of different fault severities and their fault onsets, which a real-world benchmark cannot provide. 
The second case study utilizes a benchmark dataset from an industrial multiphase flow facility~\cite{stief2019pronto}. This dataset contains artificially induced faults that introduce changes in functional relationships within the system. It also includes various operating conditions, which allows us to assess the model's robustness against novel operating conditions.
We detail the description of each case study in the following sections.}

\subsection{Case Study 1: Synthetic Dataset}
We generated a synthetic dataset mimicking an interconnected system of control variables and measurement variables to study cause-and-effect relationships in this system, with the ground truth fault severity and fault onset available.

\subsubsection{Data Generation}
\label{sec:app_syn_gen}

We simulated a system with two sinusoidal control signals influencing five measurement signals, replicating complex sensor data through nonlinear trigonometric relationships. Gaussian noise with a signal-to-noise ratio of 35 ($\text{SNR}=35$) is added to approximate realistic conditions. Faults are introduced by modifying the input-output relationships at randomly selected data segments, reflecting potential real-world system-level faults. The time of the first point in the data segment is considered as the onset of the fault.
\subsubsection{Introducing Fault Severity Levels}
To evaluate model sensitivity to faults of varying severities in particular to early subtle faults, we modulate the input-output relationship with scaling factors from 0.5 to 2.0, reflecting a range of fault severities. 
The scaling factors modulate the interdependencies between signals in our system model, selectively impacting a subset of measurements, and thereby modifying the overall dynamics of related signals. A scaling factor of 1 implies unchanged system dynamics, representing the standard operational state. Deviations from this value indicate increasingly significant changes in system dynamics, thereby increasing the fault severity. The more the scaling factor diverges from 1, the easier it becomes to detect faults, as differences in system dynamics become increasingly pronounced compared to normal behavior. 
\revtextiot{Ratios of fault samples of each fault severity class are listed in Tab.~\ref{tab:fault_rate_toy}}. The impact of fault severity on system behavior and detection complexity is thoroughly evaluated and discussed in Sec.~\ref{sec:res_toy}.
\input{tabs/anomaly_rate}

\subsection{Case Study 2: Industrial Dataset (Pronto)}

The Pronto dataset \cite{stief2019pronto} offers a benchmark for a multiphase flow facility, featuring various process variables such as pressures and flow rates. In our analysis, we selected 17 process variables sampled at a 1 Hz sampling rate, \revtextiot{as detailed in Table~\ref{tab:pronto_data}}. Due to inconsistencies in the dataset, specifically on test day five, we omitted two variables: water tank level (Ll101) and input water density (FT102-D).
Based on the system process scheme, we identify input air flow rate (FT305/302) and input water flow rate (FT102/104) as control variables, as well as input air temperature (FT305-T) and input water temperature (FT102-T) as external variables influencing the operating conditions, based on the dataset's description. 

\begin{figure}[tbhp]
\centering
    \centering
    \subfloat[t-SNE embedding space]{
        \includegraphics[width=.46\linewidth,valign=t]{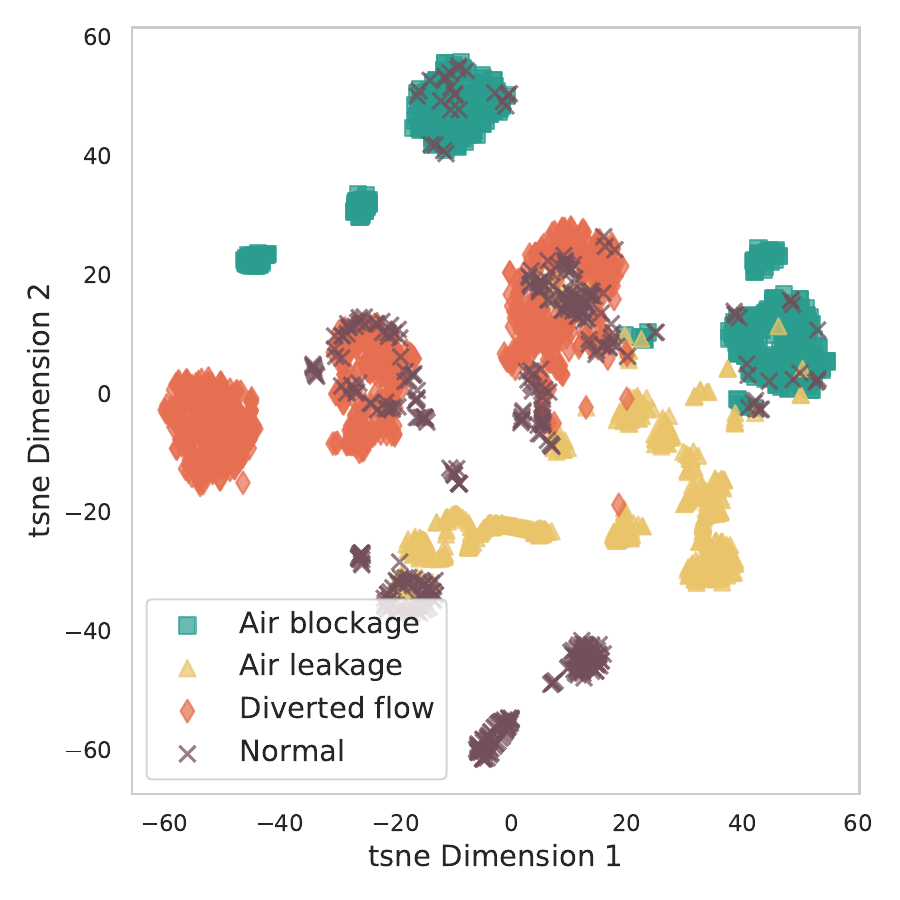}
        \label{fig:pronto_tsne}
    }
    \hfill
    \subfloat[Violin plot of PIC501 normalized by width.]{
        \includegraphics[width=.48\linewidth,valign=t]{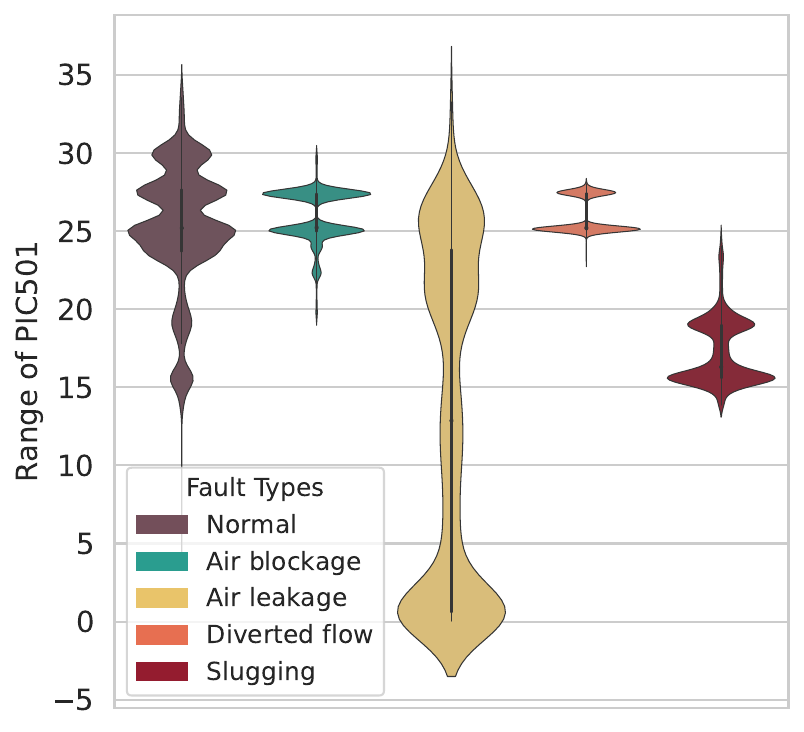}
        \label{fig:pronto_dist}
    }
    \caption{Pronto Dataset Fault Class Statistics: (a) t-SNE embedding space illustrating normal and faulty raw sequence data under two flow conditions. 
    (b) Violin plot showing the density distribution of the process variable air outlet valve 3-phase separator (PIC501).}
    \label{fig:pronto_stats}
\end{figure}

\textbf{Operating conditions}. The dataset contains 20 distinct flow conditions, defined by varying the input rates of air and water. These conditions can be divided into two major flow regimes: stable, referred to as \textit{normal}, and unstable, identified as \textit{slugging}. 
The slugging regime contains eight unique flow conditions whereas the normal regime spreads over 12 flow conditions.
In the training phase, only data from normal operating conditions is used. In the test phase, data from the slugging regime is used to evaluate whether the model is robust under new operating conditions.
Fig.~\ref{fig:pronto_dist} illustrates the data distribution exemplarily for the process variable PIC501 (Air outlet valve opening degree in the 3-phase separator). A shift occurs during the slugging condition, which indicates a transition in the system's behavior, demonstrating the challenge of distinguishing faults from novel operating conditions.  

\textbf{Fault types}. The dataset contains three types of faults—\textit{air leakage}, \textit{air blockage}, and \textit{diverted flow}, induced under two specific flow conditions within the normal flow regime. For comprehensive fault descriptions, please refer to \cite{stief2019pronto}. \revtextiot{Ratios of fault samples of each fault type are listed in Tab.~\ref{tab:fault_rate_pronto}.}
Figure~\ref{fig:pronto_tsne} presents a t-SNE visualization of the dataset, highlighting the differences in fault detection difficulties among the various fault types.
The visualization indicates that \textit{air leakage} is comparatively simpler to identify, as it forms a well-defined, separate cluster. This distinct clustering suggests a significant dissimilarity from normal conditions. In contrast, \textit{air blockage} and \textit{diverted flow} appear more challenging to detect due to the partial overlap of samples from these two fault types with the cluster representing \textit{normal} conditions. 
Furthermore, as shown in Figure~\ref{fig:pronto_dist}, the air outlet valve process variable PIC501 exceeds its normal range between 5 and 35 \% during air leakage into a negative range. Therefore, detecting air leakage can be seen as a ``point anomaly'', making it easier to detect.

\input{tabs/pronto/pronto_data}

%% file: tabs/anomaly_rate.tex
\begin{table}[tbhp]

{\color{revcolor} 
\caption{Variation of Ratios of Fault Samples ($\alpha$) Across Different Scaling Factors in the Synthetic Dataset}
\label{tab:fault_rate_toy}
\centering
\begin{tabular}{lcccccccc}
\toprule
& 0.5   & 0.75  & 0.9   & 0.95  & 1.05 & 1.1  & 1.5   & 2.0   \\
\midrule
\( \alpha \) & 0.354 & 0.407 & 0.301 & 0.302 & 0.334 & 0.378 & 0.362 & 0.425 \\
\bottomrule
\end{tabular}
}
\end{table}

\begin{table}[tbhp]
{\color{revcolor} 
\caption{Ratios of Fault Samples ($\alpha$) for Different Types of Anomalies in the Pronto Dataset}
\label{tab:fault_rate_pronto}
\centering
\begin{tabular}{lcccc}
\toprule
             & Air Leakage & Air Blockage & Diverted Flow & Slugging \\
\midrule
\( \alpha \) & 0.596       & 0.613        & 0.673     & 0.463    \\
\bottomrule
\end{tabular}
}
\end{table}

%% file: tabs/pronto/pronto_data.tex
\begin{table}[tbhp]
{\color{revcolor} 
\centering
\caption{Process variables in Pronto dataset used for fault detection~\cite{stief2019pronto}. Dark gray highlights Control Variables (CV) and light gray External Factors (EF).}
\label{tab:pronto_data}
\begin{tabular}{lll} \toprule
\textbf{Tag} & \textbf{Description} & \textbf{Unit} \\ \midrule
\rowcolor[HTML]{C0C0C0} 
 FT305/302 & Input air flow rate & $\mathrm{S} \mathrm{m}^3 \mathrm{~h}-1$ \\
\rowcolor[HTML]{EFEFEF} 
 FT305-T & Input air temperature & ${ }^{\circ} \mathrm{C}$ \\
 PT312 & Air delivery pressure & $\operatorname{bar}(g)$ \\
 \rowcolor[HTML]{C0C0C0} 
 FT102/104 & Input water flow rate & $\mathrm{kg} \mathrm{s}-1$ \\
\rowcolor[HTML]{EFEFEF} 
 FT102-T & Input water temperature & ${ }^{\circ} \mathrm{C}$ \\
 PT417 & Pressure in the mixing zone & $\operatorname{bar}(g)$ \\
 PT408 & Pressure at the riser top & $\operatorname{bar}(g)$ \\
 PT403 & Pressure in the 2-phase separator & $\operatorname{bar}(g)$ \\
 FT404 & 2-phase separator output air flow rate & $m^3 h^{-1}$ \\
 FT406 & 2-phase separator output water flow rate & $\mathrm{kg} \mathrm{s}^{-1}$ \\
 PT501 & Pressure in the 3-phase separator & $\operatorname{bar}(g)$ \\
 PIC501 & Air outlet valve 3-phase separator & $(\%)$ \\
 LI502 & Water level 3-phase separator & (\%) \\
 $\mathrm{LISO} 3$ & Water coalescer level & $(\%)$ \\
 LVC502 & Water coalescer outlet valve & $(\%)$ \\
\bottomrule
\end{tabular}
}
\end{table}


%% file: chaps/04-doe/04-experiment.tex
\section{Design of Experiments}
\label{sec:exp_design}
\revtextiot{
In this section, we detail our experimental design to assess DyEdgeGAT's fault detection performance. Specifically, we introduce the baseline methods and their configurations in Sec.~\ref{sec:doe_baselines}, evaluation metrics in Sec.~\ref{sec:doe_evaluation}, training setups in Sec.~\ref{sec:doe_training} and experiment setups in Sec.~\ref{sec:doe_hardware}.}

\input{chaps/04-doe/04-2-baselines}

\input{chaps/04-doe/04-3-metric}

\subsection{Training}
\label{sec:doe_training}
All models were trained using the Adam optimizer with a learning rate of 1e-3, the training continued for a maximum of 300 epochs with early stopping at 150 epochs with a patience of 20 steps. Experiments were repeated 5 times with different initializations, and their mean and standard deviation were reported.
In the first case study, conducted on a small synthetic dataset, a batch size of 64 and L1 loss were used, and no data normalization was needed as the generated data ranged from -1 to 1.
For the second case study, conducted on a larger dataset, a batch size of 256 was employed along with data standardization. Due to substantial differences in the statistical characteristics of the data distribution in the two case studies, L2 loss was used. 
Additionally, ReduceLROnPlateau scheduling was applied to decrease the learning rate by a factor of 0.9 after 10 consecutive epochs of non-improving validation loss.

\subsection{Experimental Setup}
\label{sec:doe_hardware}
Our proposed method, its variants, and the baseline methods were all implemented using PyTorch 1.12.1~\cite{paszke2019pytorch} with CUDA 12.0 and the PyTorch Geometric 2.2.0~\cite{fey2019fast}. 
For the synthetic dataset, the computations were performed on a server equipped with 4 NVIDIA RTX2080Ti graphic cards. 
We used neptune.ai to track the experiments.
\revtext{For the industrial dataset, computations were performed on a GPU cluster equipped with NVIDIA A100 80GB GPUs.}

%% file: chaps/04-doe/04-2-baselines.tex
\subsection{Baselines}
\label{sec:doe_baselines}
\input{tabs/baselines_tab}
\revtextrewrite{
For a comprehensive evaluation, our proposed method is benchmarked against simple and state-of-the-art fault and anomaly detection methods, detailed in Table~\ref{tab:baseline_methods}.  
These methods are classified based on their approach to modeling multivariate time series (MTS): \textit{relationship-focused} methods like Feedforward Neural Networks (FNN) and AutoEncoders (AE), \textit{dynamics-focused} methods including Long Short-Term Memory (LSTM), LSTM-based AutoEncoders (LSTM-AE), and UnSupervised Anomaly Detection (USAD), and \textit{graph-based} methods (GNNs).
}
\revtext{
In the \textit{graph-based} category, our comparison focuses on GNN-based AD methods tailored for IIoT, specifically those that simultaneously infer graphs from MTS.  
The focus on AD stems from the lack of unsupervised FD methods based on GNNs in the IIoT context, to the best of our knowledge. 
We selected Graph Deviation Network (GDN) and Multivariate Time-series Anomaly Detection with Graph Attention Network (MTAD-GAT)  because they employ attention mechanisms in graph learning, which are similar to those proposed in our approach. Additionally, we included Graph Representation Learning for Anomaly Detection (GRELEN) due to its use of graph discrepancy, aligning with our approach in anomaly score construction.

}


\revtextiot{

\input{tabs/model_hyper}
Hyperparameter tuning was tailored to each dataset. Model selection was based on the validation loss. 
For anomaly score calculation, we employ GDN's scoring function across all baselines, except for USAD which retains its dual-reconstruction score. Scores are the normalized residuals between the observed and predicted values, with normalization parameters drawn from the validation set's median and interquartile range. We employed the mean aggregation instead of the original max aggregation to reflect system relationship changes better and apply time-averaging of scores for LSTM-AE, USAD, GRELEN, and MTADGAT.

Hyperparameters were fine-tuned using grid search across plausible values, as detailed in Tab.~\ref{table:hyperparameters}, with optimal settings indicated within parentheses. The first and second values correspond to the first and second case studies, respectively. $^*$ denotes a reference to the text description for further details. Selections were based on the best validation loss.
We varied the number of layers (L) and the dimensions of the hidden layers (H) and tested different normalization techniques, such as Batch Normalization (BN) and Layer Normalization (LN). Table~\ref{table:hyperparameters} also details model-specific parameters and optimal model sizes. For FNN and AE models, the input sequence window size is 1, while for all other models, it is consistently 15 in both case studies.
\begin{itemize}
    \item \textbf{FNN}:
    We varied the number of layers and their hidden dimensions. The same independent variables used for dyEdgeGAT serve as input, mapped to the measurement variables.
    \item \textbf{AE}:
    The latent dimension of AE is the size of system-independent variables. The optimal configuration was 20-10-5-2 for the synthetic study and 20-20-20-10-10-4 for the industrial case study.
    \item \textbf{USAD}:
    We adhered to the original structure but varied the AE's latent dimension, initial training warm-up epochs, and final activation function. Adding batch normalization layers was tested but found non-beneficial.
    \item \textbf{GDN}:
    We preserved most of the original hyperparameters from the paper, as modifications had minimal impact on results. Nonetheless, the model showed sensitivity to the decoder hidden layer dimension (tested from 100 to 200) and sensor embedding dimension (tested from 20 to 50). We experimented with varying top k values for graph sparsification and discovered that the GDN model achieved optimal performance as a fully connected graph on our task.
    \item \textbf{MTADGAT}:
    The attention module was upgraded to GATv2 for its enhanced expressivity. We tested the hidden dimensions for feature and temporal attention embeddings between 10 and 40, and for the reconstruction and forecasting modules, we varied them from 10 to 80.
    \item \textbf{GRELEN}:
    We adjusted the number of graphs from 1 to 4, adhering to similar graph distribution priors as in the original paper. The hidden dimension was varied between 10 and 40. Additionally, we evaluated different anomaly scoring methods for GRELEN, identifying that the point adjustment in the original implementation was not rigorous for evaluation, further detailed in Section~\ref{sec:res_toy}.
    \item \textbf{DyEdgeGAT}:
    We varied the hidden node dimension of the node between 10 and 20, the hidden edge temporal embedding dimension between 20, 40, and 100, hidden graph dimension between 20 and 40. The dimension of the temporal encoding is between 5 and 10. 
\end{itemize}
}

%% file: tabs/baselines_tab.tex
\begin{table*}[tbh]
\centering
\caption{Summary of baseline methods for multivariate time series anomaly detection and their model input and output.}\label{tab:baseline_methods}
\begin{tabular}{p{0.1mm}p{1.8cm}p{11cm}p{1.5cm}p{1.5cm}}
\toprule
& \textbf{Method} & \textbf{Description} & \textbf{Input} & \textbf{Output} \\
\midrule
\multicolumn{3}{l}{\textit{Relationship-focused}} \\
& FNN & A Feed-Forward Neural Network (FNN) based anomaly detection method that maps control variables to measurement variables, providing a basic model of the system’s internal relationships. & $\mathbf{U}^t$ & $\mathbf{X}^t$  \\
& AE~\cite{aggarwaloutlier} & An Autoencoders (AE), similar to FNN, attempts to capture interdependencies within the data by reconstructing input data and detecting anomalies based on higher reconstruction errors & $\left[\mathbf{U}^t\parallel\mathbf{X}^t\right]$ & $\left[\mathbf{U}^t\parallel\mathbf{X}^t\right]$ \\
\midrule
\multicolumn{3}{l}{\textit{Dynamics-focused}} \\
& LSTM\cite{malhotra2015lstmad} & A Long Short-Term Memory Network mainly captures temporal patterns and weakly the inter-dependencies in MTS and identifies anomalies via deviations from predicted values. & $\left[ \mathbf{U} \parallel \mathbf{X} \right]^{t_w:t}$ & $\left[\mathbf{U} \parallel \mathbf{X}]\right]^{t+1}$\\
& LSTM-AE~\cite{malhotra2016encdecad} & An LSTM-based encoder-decoder architecture seeks to capture both temporal and inter-dependencies in the MTS. & $\left[ \mathbf{U} \parallel \mathbf{X} \right]^{t_w:t}$ & $\left[ \mathbf{U} \parallel \mathbf{X} \right]^{t_w:t}$ \\
& \revtext{USAD~\cite{audibert2020usad}} &  Unsupervised \revtext{Anomaly Detection (USAD) consists of two AEs trained adversarially on the MTS data. The output of the first AE is processed by the second AE, aiming to amplify detected anomalies.} & $\left[ \mathbf{U} \parallel \mathbf{X} \right]^{t_w:t}$ & $\left[ \mathbf{U} \parallel \mathbf{X} \right]^{t_w:t}$ \\
\midrule
\multicolumn{3}{l}{\textit{Graph-based}} \\
& GDN~\cite{deng2021gdn} & Graph Deviation Network (GDN) constructs a static graph based on the cosine similarity of the training batch data, using sensor embedding to distinguish sensor types, and employing an attention mechanism for forecasting. & $\left[ \mathbf{U} \parallel \mathbf{X} \right]^{t_w:t}$ & $\left[\mathbf{U} \parallel \mathbf{X}]\right]^{t+1}$ \\
& MTADGAT~\cite{zhao2020mtadgat} & Multivariate Time-series Anomaly Detection via Graph Attention Network (GAT) combines feature-oriented GAT\cite{velivckovic2017gat} and time-oriented GAT to handle spatial dependencies and temporal dependencies simultaneously. By performing forecasting and reconstruction concurrently, MTADGAT can model complex relationships and dynamics. & $\left[ \mathbf{U} \parallel \mathbf{X} \right]^{t_w:t}$ & $\left[ \mathbf{U} \parallel \mathbf{X} \right]^{t_w:t}$, $\left[\mathbf{U} \parallel \mathbf{X}]\right]^{t+1}$ \\
& \revtext{GRELEN~\cite{zhang22grelen}} & Graph \revtext{Relational Learning Network (GRELEN) utilizes a variational AE-based reconstruction module to dynamically infer graph structures and is the first approach to leverage learned graphs to detect anomalies from the relational discrepancy}. & $\left[ \mathbf{U} \parallel \mathbf{X} \right]^{t_w:t}$ & $\left[\mathbf{U} \parallel \mathbf{X}]\right]^{t_w+1:t}$ \\
\bottomrule
\end{tabular}
\end{table*}

%% file: tabs/model_hyper.tex
\begin{table*}[htbp]
\centering
{\color{revcolor}
\caption{Range of Hyperparameters for the applied models. The optimal hyperparameters are indicated within parentheses.}
\label{table:hyperparameters}
\begin{tabular}{lcccccc}
\toprule
\textbf{Model} & \textbf{L} & \textbf{H} & \textbf{Norm.} & \textbf{Additional Parameters} & \textbf{\revtext{No. Params}} \\ \midrule
FNN & 3-5 ($3\mid7$) & 10-50 ($10\mid 50$) & BN & first hid. dim. 20-50 ($20\mid 50$) & $515 \mid 16811$ \\
AE & 3-$5^*$ & 5-$20^*$ & BN & - & $532 \mid 3319$ \\
LSTM & 1-2 ($1\mid2$) & 10-40 ($20\mid40$) & LN & - & $2507 \mid 22935$ \\
LSTM-AE & 1 & 10-40 ($10\mid40$) & BN/LN (BN) & - & $1737 \mid 22935$ \\
\revtext{USAD} & 2-3 (2) & 10-20-30 ($10\mid30$) & BN/- (-) & warmup epochs 30-50 (50),  final act. func. (tanh$\mid$sigmoid)& $21670 \mid 75888$   \\
MTADGAT & 1-2 (1) & 10-80 ($20\mid10$) & BN & att. embed. dim. 10-40 ($20\mid10$) & $8582 \mid 5790$ \\
GDN & - & 100-200 (200) & BN & sensor embed. dim. 20-50 ($20\mid50$) & $5421 \mid 12160$  \\
\revtext{GRELEN} & - & 10-40 ($10\mid40$) & BN & num. graphs 2-4 ($2$), dist. prior ($(0.99, 0.01) \mid (0.97, 0.03)$) & $12181 \mid 63601$ \\
\revtext{DyEdgeGAT} & 2 & 10-20 ($10\mid20$) & LN+BN & temp. embed. ($5\mid10$), edge embed. ($20\mid100$), gnn embed. ($20\mid40$) & $3921 \mid 15921 $ \\
\bottomrule
\end{tabular}
}
\end{table*}

%% file: chaps/04-doe/04-3-metric.tex
\subsection{Evaluation Metrics}
\label{sec:doe_evaluation}
\revtextiot{For model performance assessment, we utilize a comprehensive set of metrics: AUC, F1 score, best F1, and best Detection Delay. In addition to these established metrics, we propose a novel metric to evaluate the model's ability in distinguishing novel operating conditions from faults. 
The metrics employed are detailed as follows:}
\begin{itemize}
    \item \textbf{AUC}: The area under the Receiver Operating Characteristic (ROC) curve, reflects discrimination capability over varying thresholds.
\item \textbf{F1}: Average of precision and recall, determined by a threshold at the 95th percentile of normal validation anomaly scores.
    \item \textbf{Best F1} (F1$^*$): The maximal F1 score obtained from the precision-recall curve.
    \item \textbf{Best Detection Delay} (Delay$^*$): Measures the time taken to identify faults after their occurrence using the threshold of best F1.
    \revtextiot{\item \textbf{Ambiguity Metric (Ambiguity)}: A novel metric defined as \( \text{Ambiguity} = 1 - 2 \cdot \left| \text{AUC} - 0.5 \right| \). It quantifies the model's inability to differentiate between normal operations and novel conditions, minimizing overfitting to specific patterns. A high ambiguity score indicates that the model effectively avoids mistaking novel operating conditions for faults. In contrast, a low score suggests that the model may struggle with this distinction, potentially leading to high false positive alarms in identifying faults. }
\end{itemize}
\revtextiot{AUC, best F1, and Delay$^*$, ambiguity are all threshold-independent and enable unbiased evaluation for any thresholding methods. 
Specifically, the ambiguity metric is applied exclusively to the industrial dataset, as detailed in Section~\ref{sec:pronto_novel_oc}. This is due to the unique relevance of the ambiguity metric in contexts where novel operating conditions are present, which is not the case for the synthetic dataset. 
}

%% file: chaps/05-results/05-results.tex
\section{Results and Discussions}
\label{sec:results}
\revtextiot{
This section evaluates the model's performance for the case studies outlined in Sec.~\ref{sec:case_study}. 
The baseline methods, their configurations, the evaluation metrics, and training setups were introduced in Sec.~\ref{sec:exp_design}.
Sec.~\ref{sec:res_toy} analyzes the model's performance on the synthetic dataset across various severity levels. 
In contrast, Sec.~\ref{sec:res_pronto} focuses on the industrial dataset, examining performance across different fault types and the ability to differentiate novel operating conditions from faults (detailed in Sec.~\ref{sec:pronto_novel_oc}). 
Additionally, Sec.~\ref{sec:res_ablation_and_sensititiy} conducts an in-depth examination of the DyEdgeGAT model, including an ablation study (Sec.~\ref{sec:res_ablation}), sensitivity analysis to sliding window size variations (Sec.~\ref{sec:res_sensitivity}), and the effect of separating system-independent variables (Sec.~\ref{sec:pronto_vartype}).
}
\input{chaps/05-results/05-1-1-toy}

\input{chaps/05-results/05-2-1-pronto}
\input{chaps/05-results/05-2-3-pronto-novel_op}
\input{chaps/05-results/05-1-2-toy-ablation}
\input{chaps/05-results/05-1-3-toy-sensitivity}
\input{chaps/05-results/05-3-sensitivity}

%% file: chaps/05-results/05-1-1-toy.tex
\subsection{Case Study I: Results on the Synthetic Dataset}
\label{sec:res_toy}

\begin{figure}[bthp]
\centering
\includegraphics[width=.85\linewidth]{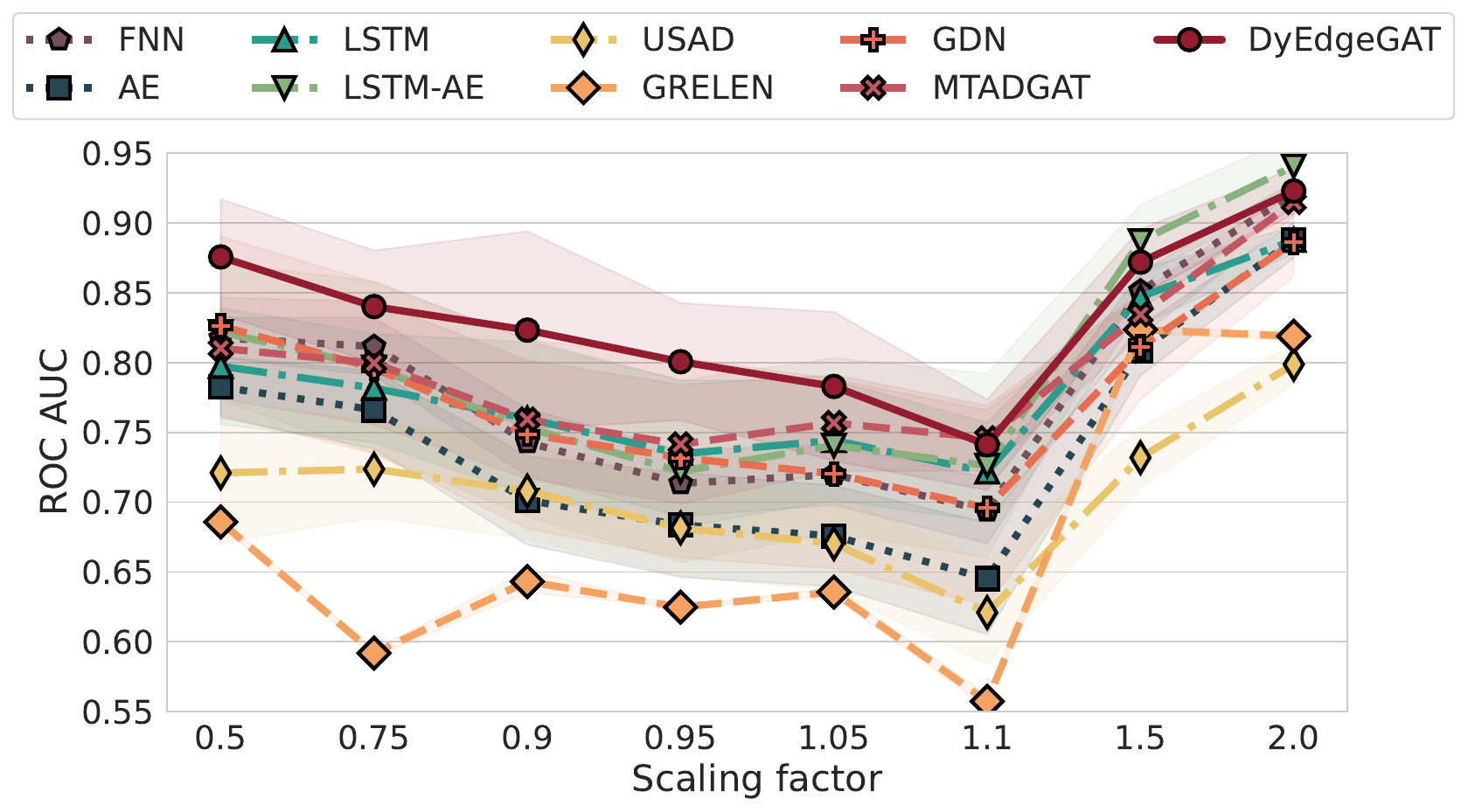}%
\\
\hspace{-5mm}
\includegraphics[width=.78\linewidth]{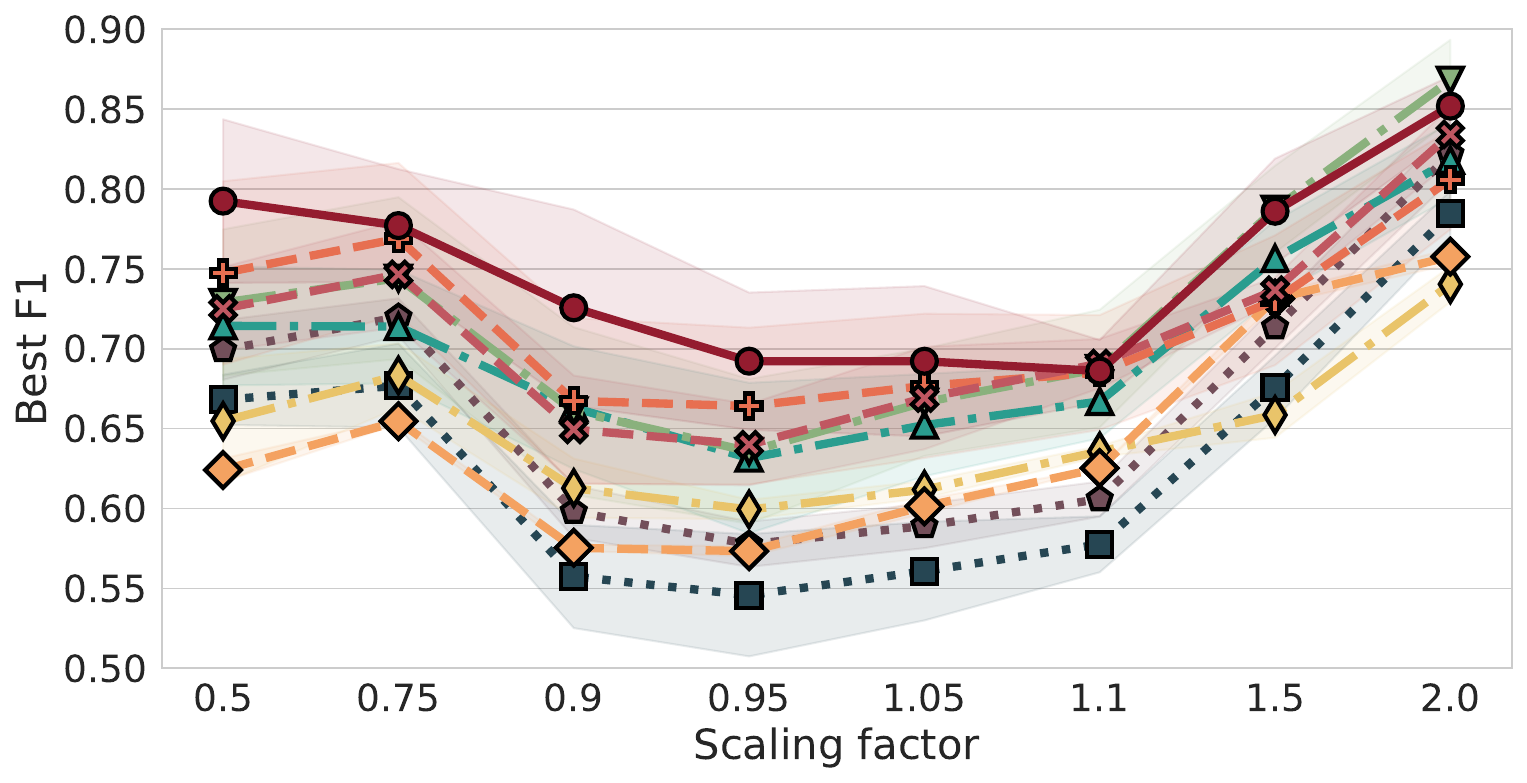}%
\caption{Comparison of model performance across different scaling factors on the synthetic dataset. A scaling factor closer to 1 indicates lower fault severity. The shaded areas around the performance lines indicate variance in model performance across multiple runs.}
\label{fig:toy_severity}
\end{figure}
\input{tabs/toy/syn_res}

\revtextiot{
\textbf{Performance across severity levels.}} Fig.~\ref{fig:toy_severity} presents the performance of all compared algorithms in terms of the AUC and best F1 score across varying scaling factors. The results demonstrate that DyEdgeGAT has consistently high AUC and F1 scores across all scaling factors, particularly at factors below 1, where the system response is damped and the fault is more challenging to detect. All other compared methods can only effectively detect faults in high fault severity scenarios where the faults have manifested significantly in the measurement signals, leading to a substantial deviation from the signal's faulty state to its normal state. These results highlight that it is essential to capture the time-evolving relationships in order to capture subtle relationship shifts and enable early fault detection.
In scenarios where scaling factors are \textbf{approximately 1}, implying minimal changes in system dynamics, fault detection becomes particularly challenging. Methods that globally model relationships (such as FNN) or dynamics (such as LSTM) struggle to detect such faults. Conversely, graph-based methods, with their ability to capture pairwise relationships, are more effective in these cases. In these scenarios, DyEdgeGAT demonstrates its superiority in terms of both AUC and F1$^*$. Its performance is notably superior at scaling factors 0.9, 0.95, and 1.05, DyEdgeGAT outperforms the other models by far. In terms of AUC, MTADGAT performs as the second-best method across most severity levels, and GDN is the second-best method in terms of F1$^*$.
When the scaling factor \textbf{exceeds 1.5}, there is a significant amplification of system dynamics. This amplification facilitates detection by methods that track system-level temporal dynamics. In such a context, LSTM-AE can outperform DyEdgeGAT.
On the contrary, at scaling factors \textbf{below 0.75}, characterized by a damped system response, dynamic-focused methods such as LSTM and LSTM-AE become less effective. The reason is the reduced discrepancy between the predicted and true values compared to cases with amplified dynamics. Moreover, USAD, which aims to amplify anomalies, struggles to identify them in this scenario due to the declining nature of the signal magnitudes.

\revtextiot{
\textbf{Overall superior performance of DyEdgeGAT.}
Table~\ref{tab:syn_res} presents the aggregated results of DyEdgeGAT and the baseline models, averaged over five runs across varying fault severities.} DyEdgeGAT consistently outperforms other comparison methods in AUC, F1, and F1$^*$ scores, highlighting its efficacy in detecting functional relationship changes. 

\revtextiot{\textbf{Dynamics focus vs. relationship-focused}.} 
LSTM and LSTM-AE generally outperform AE and FNN, indicating the importance of learning dynamic system relationships. 
Notably, FNN performs better than AE, which suggests that distinguishing control from measurement variables helps the model in learning system functionalities. This observation aligns with the recent findings of Hsu \textit{et al.} \cite{hsu2023comparison}.
The comparatively poor performance of USAD can be attributed to its focus on amplifying anomalies, which is less effective for detecting relationship shifts, as the magnitude of signals in faulty conditions does not significantly deviate from the normal state.

\revtextiot{\textbf{Graph-based}.} 
A further notable observation is that other graph-based models also demonstrate strong capabilities in signal relationship modeling. Notably, MTADGAT closely follows DyEdgeGAT in terms of the AUC score, while GDN ranks second in terms of the F1 score after DyEdgeGAT. 
\revtextiot{
GRELEN's underperformance, contrary to previous findings, is due to excluding the non-rigorous point adjustment step in its original anomaly score calculation, as outlined by Kim~\textit{et al.}~\cite{kim2022towards}.
To demonstrate the significant impact of point adjustment, we report the averaged AUC score of GRELEN with and without point adjustment in Tab.~\ref{tab:syn_grelen_result}.
}

\revtextiot{\textbf{Delay detection and F1 score performance}.}
DyEdgeGAT exhibits a moderately higher {Delay}$^*$, yet its superior F1$^*$ indicates more accurate timely detection. In practical applications, it is often preferred to have accurate fault detection over those that are early but are potentially false alarms. 
Furthermore, DyEdgeGAT achieves the highest F1 score, which is noteworthy considering that it is derived from a simple and straightforward thresholding method. In this approach, any score exceeding the 95th percentile of the validation set is considered indicative of a fault. Despite its simplicity, this method proves highly effective, enhancing DyEdgeGAT's suitability for real-world applications.
\input{tabs/toy/grelen_syn_res}

%% file: tabs/toy/syn_res.tex
\begin{table}[htb]
\centering
\caption{
Model Performance comparison on the synthetic dataset with 5 runs over 8 test cases. The best metric score is highlighted in bold and the second best model in underscore.}
\label{tab:syn_res}
\begin{tabular}{lcccc}
\toprule
 & \textbf{AUC}$\uparrow$ & \textbf{F1}$\uparrow$ & \textbf{F1}$^*\uparrow$ & \textbf{Delay}$^*$ $\downarrow$ \\
\midrule
AE & 0.74 ± 0.03 & 0.56 ± 0.03 & 0.63 ± 0.02 & 30.0 ± 9.7 \\
FNN & 0.78 ± 0.02 & 0.55 ± 0.01 & 0.67 ± 0.01 & 21.7 ± 9.0 \\ \midrule
LSTM & 0.78 ± 0.04 & 0.61 ± 0.00 & 0.70 ± 0.03 & 23.3 ± 11.4 \\
LSTM-AE & \underline{0.80 ± 0.05} & 0.61 ± 0.00 & \underline{0.72 ± 0.04} & \textbf{9.2 ± 5.9} \\
USAD & 0.71 ± 0.03 & 0.61 ± 0.00 & 0.65 ± 0.01 & 43.4 ± 5.1 \\ \midrule
GDN & 0.78 ± 0.06 & \underline{0.66 ± 0.07} & 0.72 ± 0.04 & \underline{19.4 ± 10.6} \\
GRELEN & 0.67 ± 0.00 & 0.42 ± 0.01 & 0.64 ± 0.00 & 33.6 ± 3.2 \\
MTADGAT & 0.80 ± 0.05 & 0.65 ± 0.03 & 0.71 ± 0.02 & 25.0 ± 5.7 \\ \midrule
DyEdgeGAT & \textbf{0.83 ± 0.04} & \textbf{0.69 ± 0.08} & \textbf{0.75 ± 0.04} & 21.4 ± 5.6 \\
\bottomrule
\end{tabular}
\end{table}

%% file: tabs/toy/grelen_syn_res.tex
\begin{table}[tbhp]
\centering
\caption{Performance evaluation of GRELEN with and without Point Adjustment (PA) on the synthetic dataset.}
\label{tab:syn_grelen_result}
\begin{tabular}{lcc}
\toprule
 & \textbf{AUC} & \textbf{AUC\textsubscript{PA}} \\
\midrule
Degree & 0.548 ± 0.051 & 0.970 ± 0.010 \\
Node Scaled & 0.676 ± 0.005 & 0.943 ± 0.003 \\
\bottomrule
\end{tabular}
\end{table}

%% file: chaps/05-results/05-2-1-pronto.tex
\subsection{Case Study II: Results on the Industrial Dataset }
\subsubsection{Fault Detection Performance Across Various Fault Types}

\label{sec:res_pronto}
\begin{figure}[tbhp]
\centering
\includegraphics[width=.85\linewidth]{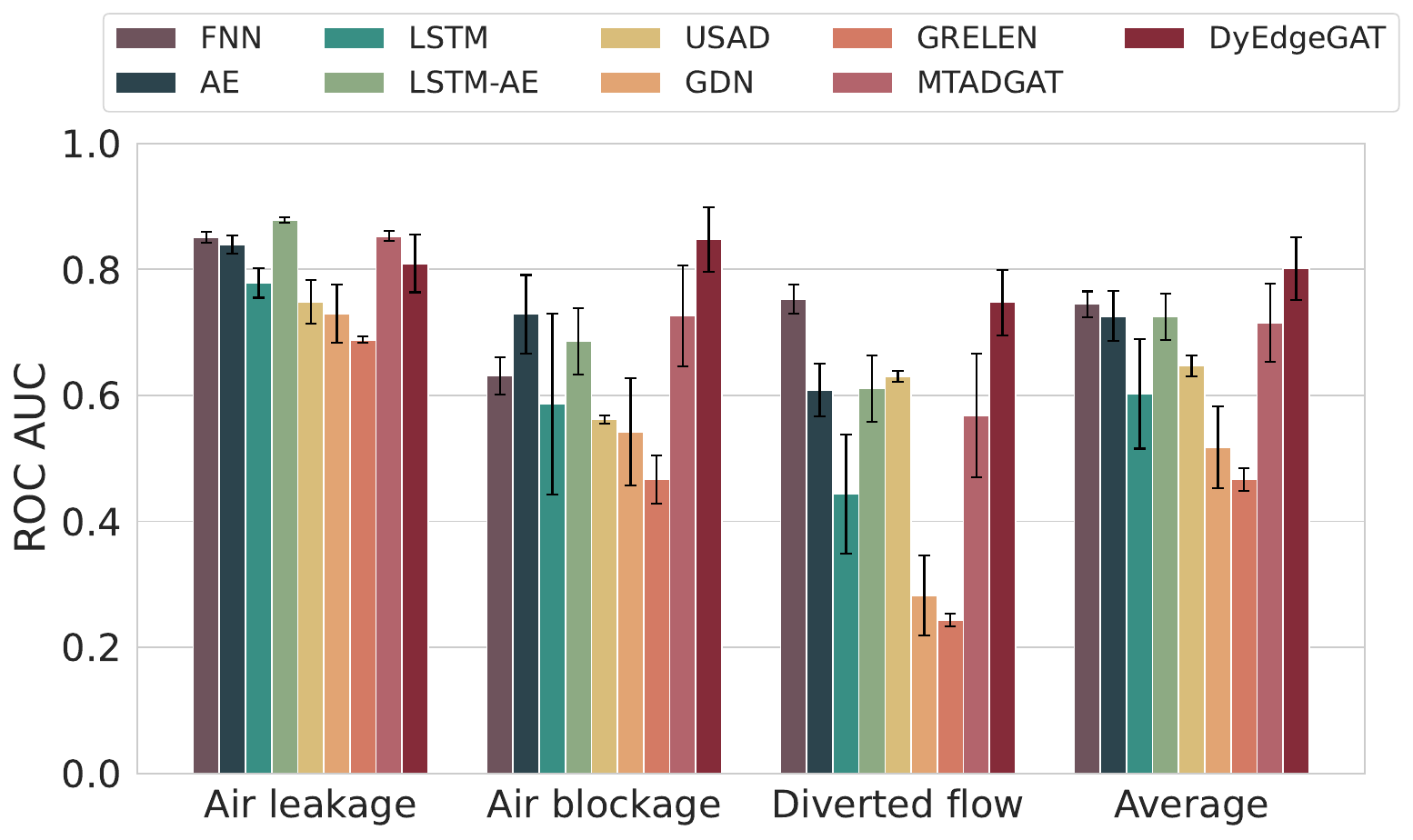}\\
\hspace{-0.4cm}
\includegraphics[width=.84\linewidth]{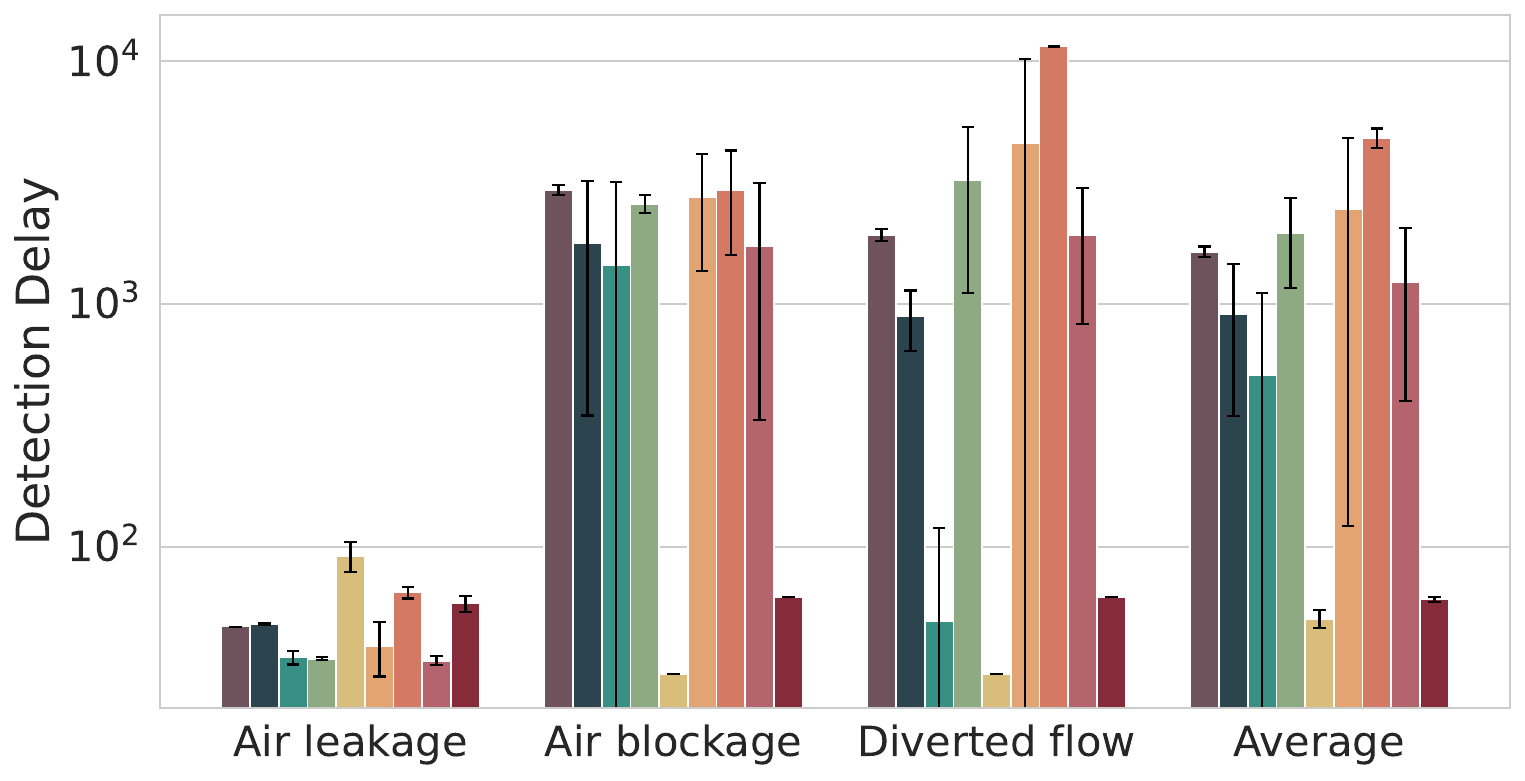}
\caption{Comparison of fault detection performance on the Pronto Dataset under different fault types.}
\label{fig:pronto-result}
\end{figure}
\input{tabs/pronto/pronto_res}

We evaluate the performance of DyEdgeGAT on the Pronto dataset on three distinct fault types: air leakage, air blockage, and diverted flow, and compare it to the baseline methods. Figure~\ref{fig:pronto-result} shows the AUC scores for each fault type, as well as the average performance across all fault types. Notably, DyEdgeGAT is the only method that demonstrates consistently strong performance across all three fault types in terms of both the AUC score and the detection delay, highlighting its superior ability to detect relationship shifts for fault detection.

\revtext{
\textbf{Air Leakage.} Characterized by cyclic behavior in the multiphase flow facility, air leakage creates a noticeable deviation from normal patterns, making its detection relatively straightforward. DyEdgeGAT, though not the best performer on this fault type, still demonstrates a competitive performance of AUC 0.81 and a detection delay of 59s.

\textbf{Air Blockage.} On this fault type, DyEdgeGAT outperforms all baselines, achieving an AUC of 0.847. This score outperforms the second-best model, AE, by a notable margin of 0.119. Furthermore, DyEdgeGAT achieves a very low detection delay of 62s, well below AE's 1777s. The nature of continuous flow in air blockages without obvious pressure drops makes it a more complex fault to detect. The poor performance of FNN with an AUC of 0.631 in this fault, despite its effectiveness in detecting other faults, indicates that it is not only important to model the functional relationships but also to track their evolution over time to enable accurate fault detection.

\textbf{Diverted Flow.}  This fault presents the most challenge with an average AUC of 0.543, with only DyEdgeGAT performing adequately in both an AUC score of 0.748 and a detection delay of 62s. Diverted flow leads to noticeable shifts in functional relationships within the system, which explains FNN's high AUC score. The inability of the dynamics-focused method to detect this type of fault suggests that the signals remain within their normal range and exhibit temporal patterns that resemble those observed under normal conditions. The drop in performance of graph-based models may be attributed to their approach of treating all signal nodes uniformly, ignoring the cause-and-effect relationships in the systems, which reveals the disadvantages of a homogeneous node treatment of all signals.}


\revtextrewrite{
To provide a broader perspective, we now evaluate the average performance of all models on all fault types with two additional metrics, F1 and F1$^*$. Tab.~\ref{tab:pronto_results} reveals that DyEdgeGAT on average outperforms all other compared models in terms of AUC, F1, and F1$^*$. Additionally, it exhibits constantly low detection delay across all fault types with a small variance, indicating its capability for timely fault detection. The following two notable observations emerge from the overall performance:}
\revtext{

\textbf{Importance of functional relationship modeling}. Firstly, models that prioritize functional relationship modeling, such as FNN and AE, demonstrate better performance than those focusing on temporal dynamics like LSTM. This suggests that the changes in functional relationships within the system's variables are more pronounced than changes in their dynamics for the Pronto dataset, where faults were incrementally introduced and the fault severity was gradually increased.

\textbf{Suboptimal performance of GNN models}. Secondly, graph-based models such as GDN and GRELEN exhibit suboptimal performance on the fault detection task across all fault types in the Pronto case study. This is likely due to their inability to distinguish between system-dependent and system-independent variables. In the Pronto dataset, this differentiation is crucial because system-independent variables remain largely unaffected by faults and should not be modeled as system-dependent variables. Without separating system-independent variables, GNN-based methods are more susceptible to mistaking novel operating conditions as faults. The impact of novel operating conditions on the model's fault detection performance is analyzed in the subsequent Sec.~\ref{sec:pronto_novel_oc}.
}

%% file: tabs/pronto/pronto_res.tex
\begin{table}[tbhp]
\centering
\caption{Performance comparison on pronto dataset with 5 runs over three fault classes. The best metric score is highlighted in bold and the second best model in underscore.}
\label{tab:pronto_results}
\begin{tabular}{lcccc}
\toprule
 & \textbf{AUC}$\uparrow$ & \textbf{F1}$\uparrow$ & \textbf{F1}$^*\uparrow$ & \textbf{Delay}$^*$ $\downarrow$ \\
\midrule
AE & 0.73 ± 0.04 & 0.40 ± 0.02 & \underline{0.83 ± 0.02} & 905 ± 559 \\
FNN & \underline{0.74 ± 0.02} & 0.63 ± 0.05 & 0.82 ± 0.01 & 1644 ± 83 \\\midrule
LSTM & 0.60 ± 0.09 & \underline{0.70 ± 0.11} & 0.80 ± 0.02 & 511 ± 599 \\
LSTM-AE & 0.73 ± 0.04 & 0.61 ± 0.04 & 0.81 ± 0.01 & 1952 ± 784 \\
USAD & 0.65 ± 0.02 & 0.27 ± 0.04 & 0.80 ± 0.00 & \textbf{51 ± 4} \\\midrule
GDN & 0.52 ± 0.06 & 0.27 ± 0.05 & 0.78 ± 0.01 & 2467 ± 2345\\
GRELEN & 0.47 ± 0.02 & 0.29 ± 0.01 & 0.77 ± 0.00 & 4838 ± 451\\
MTADGAT & 0.72 ± 0.06& 0.40 ± 0.02& 0.82 ± 0.02& 1232 ± 834\\\midrule
DyEdgeGAT & \textbf{0.80 ± 0.05} & \textbf{0.83 ± 0.02} & \textbf{0.86 ± 0.02} & \underline{61 ± 1}  
\\
\bottomrule
\end{tabular}
\end{table}

%% file: chaps/05-results/05-2-3-pronto-novel_op.tex
\subsubsection{Performance on Novel Operating Conditions}
\label{sec:pronto_novel_oc}
\begin{figure}[htbp]
\centering
\includegraphics[width=.8\linewidth]{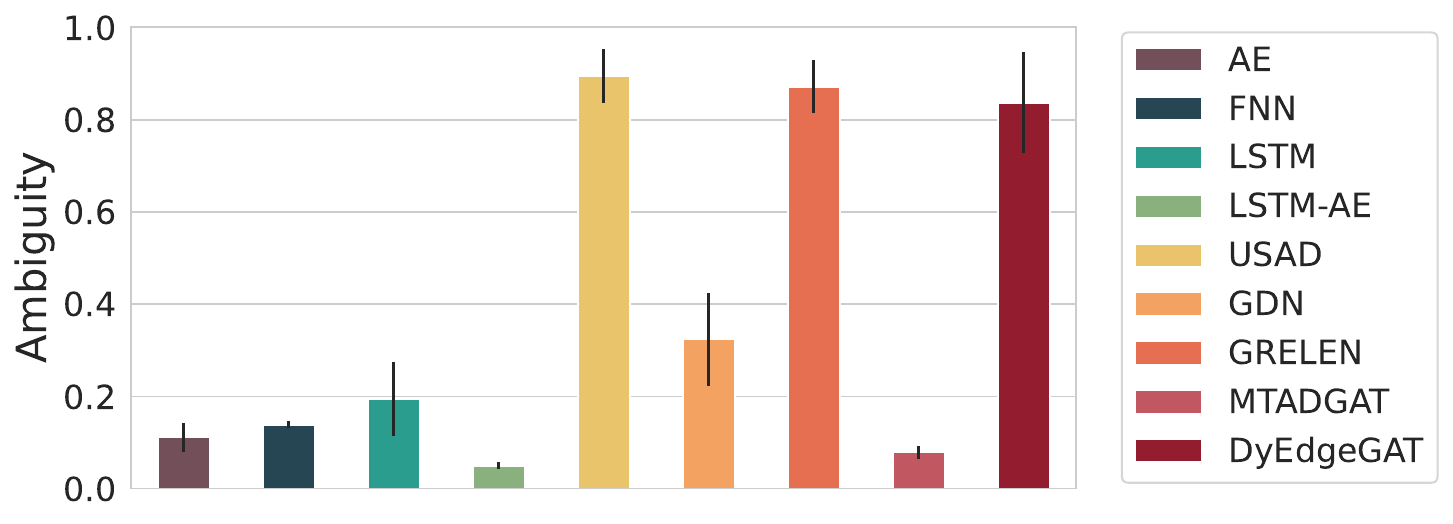}
\caption{
Comparative model evaluation on the Pronto dataset under novel operating conditions, specifically for the slugging condition. The ambiguity metric reflects the model's disability to distinguish between normal and novel operating conditions.}
\label{fig:pronto_ambiguity}
\end{figure}

\revtext{
In IIoT systems, fault detection models must be robust to novel operating conditions to reduce false alarms. 
We assess this aspect on the novel operating condition of ``slugging'' in the Pronto dataset. 
Fig.~\ref{fig:pronto_ambiguity} illustrates that our DyEdgeGAT model achieves a relatively high ambiguity score. This indicates that DyEdgeGAT demonstrates robustness and generalizability, accurately distinguishing novel operating conditions from faults.
In contrast, other models with good fault detection performance on the pronto dataset, such as AE, FNN, and LSTM-AE, have significantly lower ambiguity scores. 
A change in operating conditions can result in dynamics that are significantly different from those observed, even if the underlying functional relationships remain the same. 
This indicates that these models are highly sensitive to changes in operating conditions, leading to misclassification of new system dynamics induced by novel operating conditions as faults.
Conversely, the high ambiguity scores of USAD and GRELEN stem from their overall underperformance in fault detection. These models struggle not only to differentiate between novel operating conditions and faults but also fail to detect faults effectively. This inefficiency is particularly highlighted by their low AUC scores in identifying air blockage and diverted flow.
}

%% file: chaps/05-results/05-1-2-toy-ablation.tex
\subsection{Ablation Study and Sensitivity Analysis}
\label{sec:res_ablation_and_sensititiy}
\subsubsection{Ablation Study}
\label{sec:res_ablation}
\revtextrewrite{
To understand how each component of our proposed methodology contributes to the overall fault detection performance, we conducted an ablation study on the synthetic dataset with five setups:
\begin{itemize}
\item \textbf{w/o oc aug}: Omitting the augmentation of operating condition context, considering each node in the temporal graph equally. 
\item \textbf{w/o dyn. graph}: Removing dynamic edge construction, opting for static graph construction via MTADGAT's feature attention and GDN's top-k mechanism.
\item \textbf{w/o reverse}: \revtext{Skipping signal reversion and training the decoder to reconstruct the original signal.}
\item \textbf{w/o time}: Omitting temporal encoding, potentially weakening temporal dependency capturing.
\item \textbf{w/o topl}: Eliminating topology-based anomaly score, assigning uniform weights to all node residuals.
\end{itemize}
}
\input{tabs/toy/syn_ablation_results}
\begin{figure}[htbp]
\centering
\includegraphics[width=0.75\linewidth]{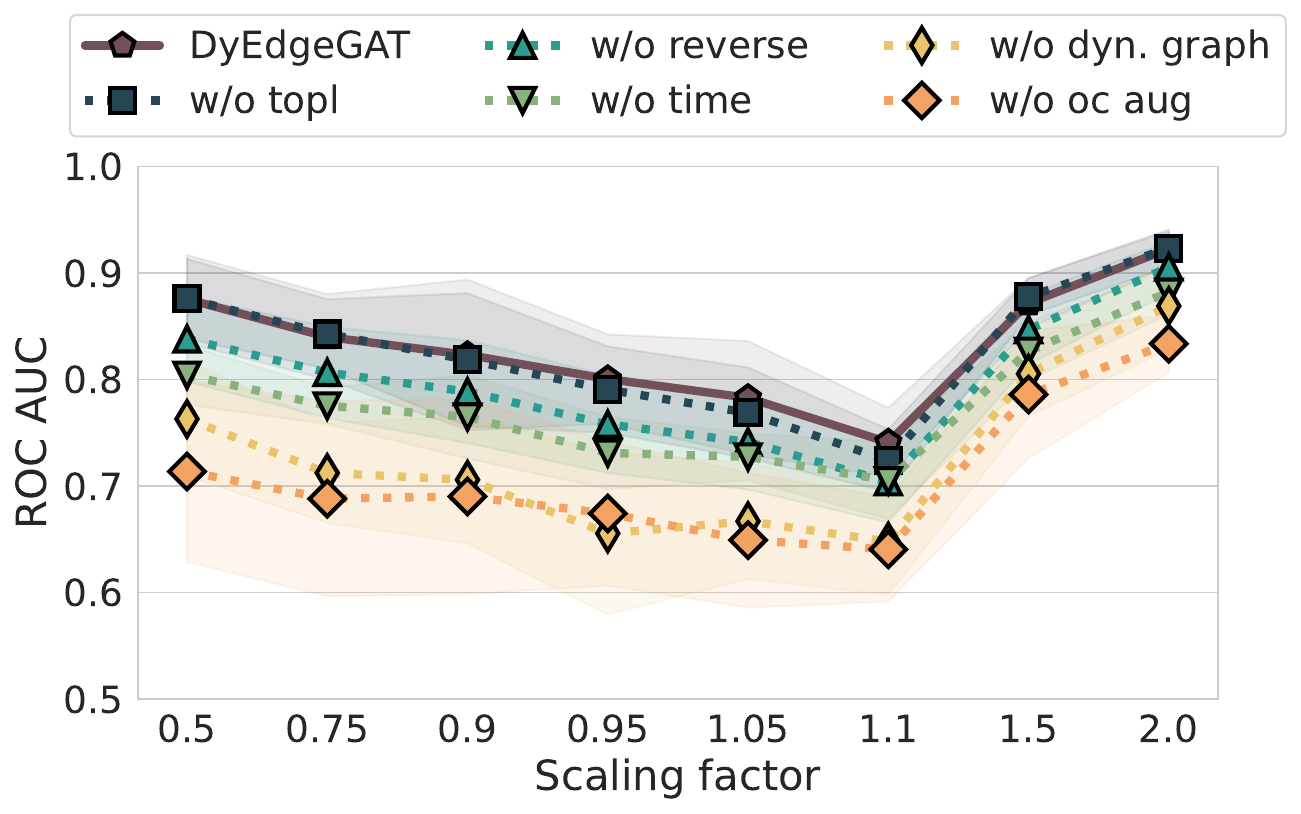}
\caption{Ablation study showing the AUC of the DyEdgeGAT model on a synthetic dataset with varying fault severity, denoted by the scaling factor. Each curve indicates the model's robustness to the exclusion of specific features, such as dynamic edge construction, operating condition context augmentation, reversed signal reconstruction, temporal encoding, and topology-based anomaly score.}
\label{fig:syn_ablation}
\end{figure}

\revtextrewrite{
The results of the ablation study, outlined in Table~\ref{tab:syn_ablation} and visualized by Fig.~\ref{fig:syn_ablation}, demonstrate a performance degradation with the removal of any component from the proposed DyEdgeGAT algorithm. Notably, the elimination of topology-based anomaly scoring, signal reversion reconstruction, and temporal encoding results in a modest reduction in performance metrics. 
The topology-based anomaly score, while generally exerting a minor influence on the model, becomes more relevant when the scaling factor approaches 1. This increase in influence can be attributed to its mechanism of normalizing anomaly scores by the strength of signal dynamics, proving valuable for detecting subtle shifts in relationships.
In contrast, excluding operating conditions (OC) context augmentation and dynamic edge construction leads to a significant decrease in both AUC and F1$^*$ scores. The impact of OC context augmentation is particularly noticeable when system dynamics undergo substantial changes, as observed at scaling factors near 0.5 or 2.0. Conversely, the importance of the proposed dynamic edge is more pronounced in detecting subtle changes in system dynamics at lower scaling factors, becoming most critical at a scaling factor of 0.95. 
}

%% file: tabs/toy/syn_ablation_results.tex
\begin{table}[htbp]
\centering
\caption{Ablation study on the synthetic dataset with 5 runs}
\label{tab:syn_ablation}
\begin{tabular}{lcc}
\toprule
 & \textbf{AUC}$\uparrow$ & \textbf{F1}$^*\uparrow$ \\
\midrule
DyEdgeGAT & \textbf{0.832 ± 0.040} & \textbf{0.750 ± 0.039} \\ 
- w/o topl & 0.828 ± 0.035 & 0.745 ± 0.034 \\
- w/o time & 0.777 ± 0.028 & 0.702 ± 0.026 \\
- w/o reverse & 0.799 ± 0.040 & 0.721 ± 0.026 \\
- w/o dyn. graph & 0.728 ± 0.053 & 0.693 ± 0.035 \\
- w/o control & 0.710 ± 0.067 & 0.666 ± 0.034 \\
\bottomrule
\end{tabular}
\end{table}

%% file: chaps/05-results/05-1-3-toy-sensitivity.tex
\subsubsection{Sensitivity Analysis of the Sliding Window Size}
\label{sec:res_sensitivity}
\begin{figure}[tbhp]
\centering
\includegraphics[width=.75\linewidth]{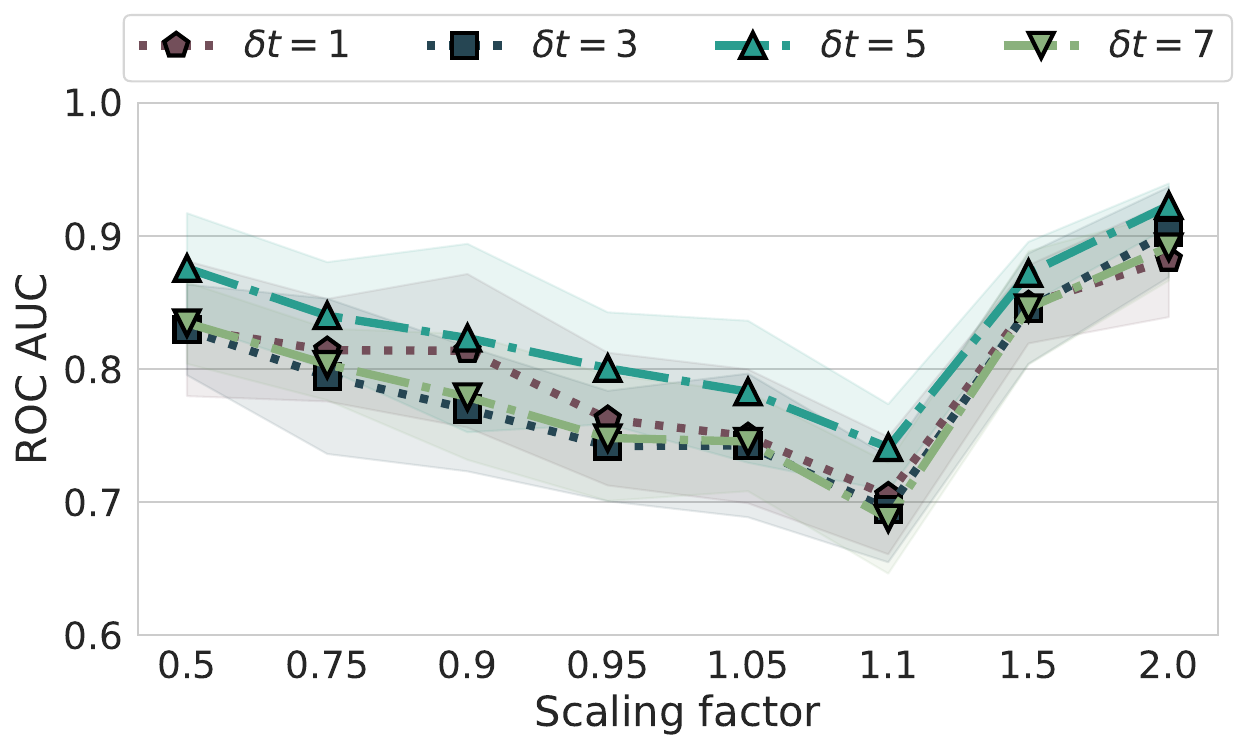}
\caption{Sensitivity analysis of DyEdgeGAT performance, measured by ROC-AUC, over a range of scaling factors for different sliding window sizes ($\delta t$). }
\label{fig:toy-sensitivity}
\end{figure}

\revtext{
The sliding window size $\delta t$ is a critical parameter in the DyEdgeGAT algorithm, influencing the construction of dynamic edges and the extraction of edge dynamic features. Our sensitivity analysis, summarized in Tab.~\ref{tab:syn_sensitivity} and illustrated in Fig.~\ref{fig:toy-sensitivity}, demonstrates the effect of $\delta t$ on the algorithm's fault detection performance. 
The analysis shows that DyEdgeGAT's performance is robust across different sliding window sizes, exhibiting competitive performance even at suboptimal choices of $\delta t$ compared to optimized baselines.

The sliding window size $\delta t$ determines the granularity of temporal information encoded into the graph structure. With an input sequence length of 15 and a GRU-based edge encoder in Eq.~\ref{eq:edge_gru}, it is essential to find a balance between local temporal resolution and the preservation of sufficient data points to capture the evolution of dynamics. A small $\delta t$ (e.g., 1) may compromise temporal resolution, whereas a too large $\delta t$ (e.g. 7) may not offer sufficient context for the GRU to capture meaningful dynamics,  potentially resulting in poorer performance.
Consequently, $\delta t=5$ emerges as the optimal sliding window size. This setting allows the model to effectively capture a comprehensive range of temporal data, ensuring both detailed temporal resolution and a thorough representation of dynamic evolution.
}
\input{tabs/toy/sensitivity_results}

%% file: tabs/toy/sensitivity_results.tex
\begin{table}[tbhp]
\centering
\caption{Sensitivity analysis of TimeGAT's performance on the synthetic dataset across different sliding window sizes.}
\label{tab:syn_sensitivity}
\begin{tabular}{ccc}
\toprule
 & \textbf{AUC}$\uparrow$ & \textbf{F1}$^*$$\uparrow$ \\
\midrule
$\delta t=1$ & 0.801 ± 0.045 & 0.726 ± 0.030 \\
$\delta t=3$ & 0.790 ± 0.044 & 0.704 ± 0.031 \\
$\delta t=5$ & \textbf{0.832 ± 0.040} & \textbf{0.750 ± 0.039} \\
$\delta t=7$ & 0.792 ± 0.037 & 0.717 ± 0.033 \\
\bottomrule
\end{tabular}
\end{table}

%% file: chaps/05-results/05-3-sensitivity.tex
\subsubsection{Impact of Separating System-Independent Variables}
\label{sec:pronto_vartype}
\begin{figure}[tbhp]
\centering
\includegraphics[width=.8\linewidth]{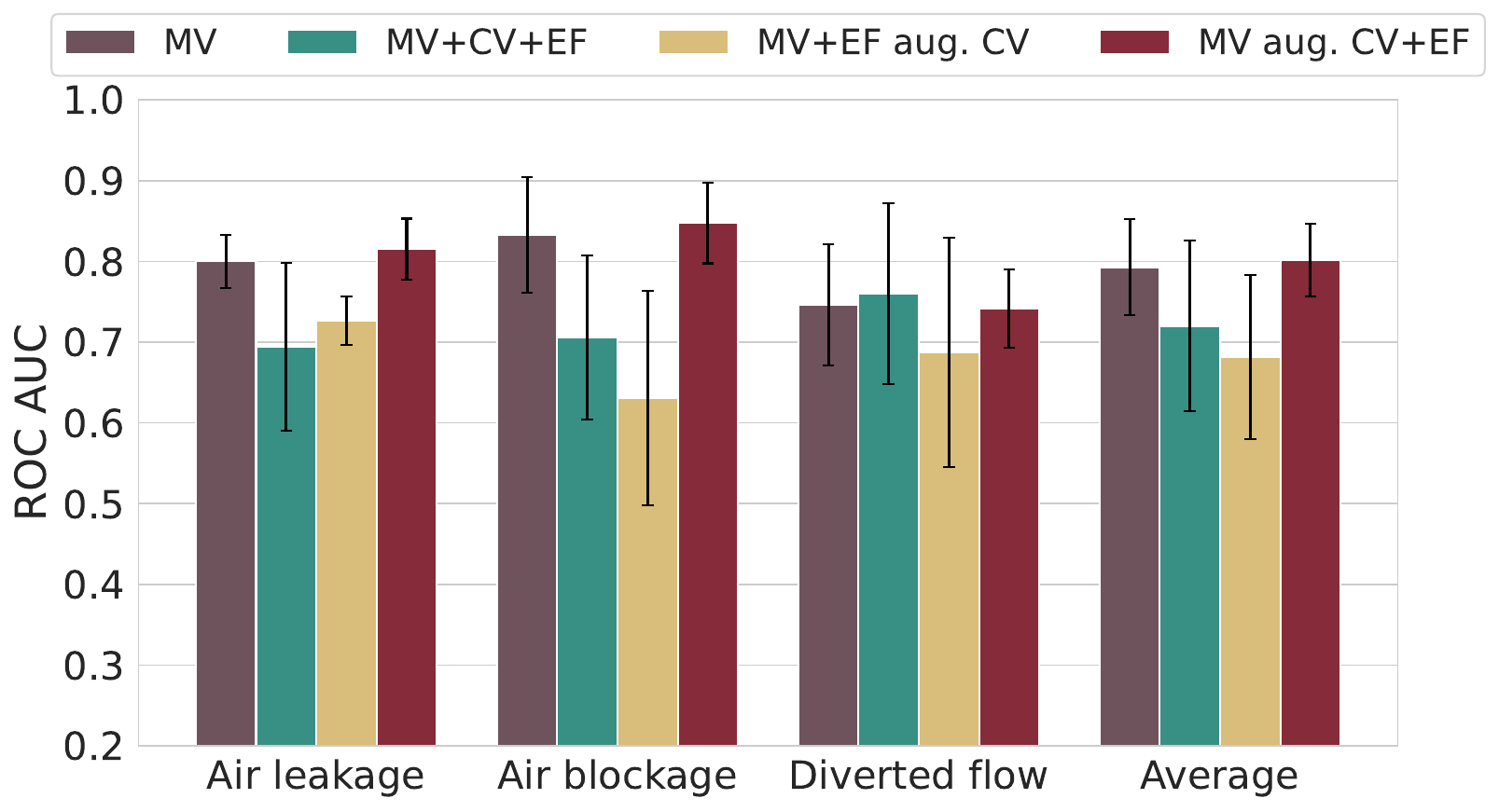}
\caption{AUC scores for DyEdgeGAT with different combinations of system-independent variables across various fault types. The variables include Measurement Variables (MV), Control Variables (CV), and External Factors (EF).}
\label{fig:pronto-vartype}
\end{figure}

\revtext{
In the pronto dataset description, only control variables are used to define an operating condition~\cite{stief2019pronto}. This ablation study aims to demonstrate the importance of incorporating external factors for a more comprehensive description of operating conditions, as reflected in the performance improvement of DyEdgeGAT shown in Fig.~\ref{fig:pronto-vartype}. 
We categorize input air and water flow rates as Control Variables (CV) and their corresponding temperatures as External Factors (EF), with the remaining variables classified as Measurement Variables (MV).
The performance of DyEdgeGAT varies depending on how system-independent variables are incorporated into the model. The best performance is achieved when CV and EF are modeled as operating condition contexts to augment the extraction of node dynamics, with MV modeled as system-dependent variables (MV aug. CV+EF). 
This highlights that CV alone is not enough to represent the operating condition. When EF is not considered as part of the operating condition but treated as a system-dependent variable, the model's performance is even worse than not augmenting the operating condition context at all.
The MV+CV+EF configuration treats all variables in the same way, as commonly employed by other graph-based methods. Even under this configuration, DyEdgeGAT outperforms all other GNN-based methods with discrete time dynamic graphs on the air blockage and diverted flow fault types, suggesting that the aggregated dynamic graph representation is superior for detecting relationship shifts.
Another notable observation is that the MV-only configuration performs on par with MV aug. CV+EF. The key distinction between these two configurations is the inclusion of operating condition context in the node dynamics extraction for the MV aug. CV+EF.
This implies that the dynamic edge module alone, even without the augmentation of operating conditions in the node dynamics, is effective in capturing temporal relationships. 
In conclusion, separating system-independent from system-independent variables is crucial for accurate fault detection. Treating them in the same way can degrade model performance even more than using only a subset of them. It is not sufficient to consider only control variables; rather, it is crucial to identify the external factors associated with them.}

%% file: chaps/06-conclusion.tex
\section{Conclusions and Future Outlook}
\label{sec:conclusion}
In this study, we propose DyEdgeGAT, an unsupervised framework for early fault detection that utilizes graph attention to dynamically construct edges. This approach effectively captures evolving relationships between MTS, enhancing early fault detection. We incorporate operating condition context into node dynamics extraction, improving thereby robustness against novel operating conditions and mitigating false alarms.
DyEdgeGAT outperforms existing discrete-time graph-based methods on both synthetic and real-world industrial datasets, particularly in detecting early faults with a low severity that are often missed by other methods. Furthermore, it is effective in distinguishing between faults and novel operating conditions, a task where state-of-the-art methods typically struggle.
Our ablation study highlights the efficacy of each component in the proposed architecture. 
Additionally, we examined the impact of sliding window size and showed the impact of separating system-independent variables to enable robust reliable fault detection. In particular, it is important to identify both control variables and external factors in order to describe the context of operating conditions.
\revtextiot{
In terms of real-world applicability, DyEdgeGAT demonstrates significant potential for straightforward and resource-efficient  deployment in industrial environments. This is attributed to its compact model size and short inference time. Such compactness not only facilitates easier integration into existing industrial systems but also ensures efficiency, which is crucial in environments with limited computational resources. 
Furthermore, DyEdgeGAT's rapid data processing capability enables timely fault detection, thus enhancing overall industrial safety and productivity. 
The model's unsupervised nature, requiring minimal labeled data, further enhances its practicality for industrial use, especially in scenarios where acquiring extensive fault data is challenging. 
While DyEdgeGAT is effective for moderately large systems, scaling it to very large systems presents challenges, particularly due to the method's quadratic complexity in pairwise dynamic edge construction. 
Future research should explore the incorporation of physical prior or hierarchical structures to introduce physical biases in graph construction and enhance scalability while maintaining detection accuracy.
Additionally, further investigation into the integration of operating state context is beneficial, especially in scenarios where information about system-independent variables is limited. 
}